\documentclass[11pt]{article}

\usepackage[final]{acl}

\usepackage{times}
\usepackage{latexsym}

\usepackage[T1]{fontenc}
\usepackage[utf8]{inputenc} 
\usepackage{microtype} 
\usepackage{inconsolata} 

\usepackage{graphicx} 

\definecolor{tablecolor}{rgb}{0.8,0.8,0.8}

\usepackage{xspace}
\usepackage{url}            
\usepackage{booktabs}       
\usepackage{amsfonts}       
\usepackage{nicefrac}       
\usepackage[most]{tcolorbox}
\usepackage{xcolor}         
\usepackage{tikz-dependency}
\usepackage{comment}
\usepackage{adjustbox}
\usepackage{array}
\usepackage{multirow}
\usepackage{tabularx}
\usepackage{caption}
\usepackage{subcaption}
\usepackage{linguex}

\usepackage{textgreek}
\usepackage{amssymb}
\usepackage{amsmath}
\usepackage{newunicodechar}
\usepackage{tipa}
\usepackage[T4,T1]{fontenc}
\usepackage{siunitx}
\usepackage{pifont}
\usepackage{ragged2e}
\usepackage{enumitem}
\usepackage{longtable}
\usepackage{color, colortbl}
\usepackage{longtable}
\usepackage{float}
\usepackage{placeins}

\definecolor{promptbg}{RGB}{240, 240, 240}
\definecolor{promptblue}{RGB}{0, 70, 127}

\newtcolorbox{promptbox}[1]{
  enhanced,
  colback=promptbg,
  colframe=promptblue,
  boxrule=0.8pt,
  arc=3pt,
  width=\columnwidth,
  top=8pt,
  bottom=8pt,
  left=8pt,
  right=8pt,
  fonttitle=\bfseries,
  coltitle=white,
  colbacktitle=promptblue,
  title=#1
}

\title{AfriSUD: A Dependency Treebank Collection for Evaluating Models on African Languages}

\author{
\textbf{Happy Buzaaba}$^{1,2}$,
\textbf{Cheikh Mouhamadou Bamba Dione}$^{3}$,
\textbf{David Ifeoluwa Adelani}$^{4,5}$, \\
\textbf{Sylvain Kahane}$^{6}$,
\textbf{Kim Gerdes}$^{7}$,
\textbf{Bruno Guillaume}$^{8,9,10,11}$,
\textbf{Shamsuddeen Hassan Muhammad}$^{14}$, \\
\textbf{Kevin Guan}$^{1}$,  
\textbf{Aremu Anuoluwapo}$^{12}$,
\textbf{Naome A. Etori}$^{13}$,
\textbf{Utitofon Inyang}$^{15}$,
\textbf{Peter Nabende}$^{16}$, \\
\textbf{David Sabiiti Bamutura}$^{18,19}$,
\textbf{Andiswa Bukula}$^{17}$, 
\textbf{Chinedu Uchechukwu}$^{21}$, \\
\textbf{Rooweither Mabuya}$^{22}$,
\textbf{Idris Akinade}$^{20}$,
\textbf{Christiane Fellbaum}$^{1}$
\\[0.5em]
\normalfont\small
$^1$Princeton University \quad
$^2$Princeton Language and Intelligence, Princeton University \quad
\small
$^3$Gaston Berger University \\
\small
$^4$Mila, McGill University \quad
$^5$Canada CIFAR AI Chair \quad
$^6$Paris Nanterre University \quad
$^7$Paris-Saclay University \quad
$^8$CNRS \\
\small
$^9$Inria \quad
$^{10}$LORIA \quad 
$^{11}$Universit\'{e} de Lorraine \quad
$^{12}$University of Trento \quad
$^{13}$University of Minnesota--Twin Cities \\
\small
$^{14}$Imperial College London, UK \quad
$^{15}$Binghamton University \quad
$^{16}$Makerere University \quad
\small
$^{17}$Penn State University \\
\small
$^{18}$Mbarara University of Science and Technology \quad
$^{19}$Chalmers University of Technology \quad
$^{20}$University of Ibadan \\
\small
$^{21}$Nnamdi Azikiwe University \quad
$^{22}$South African Centre for Digital Language Resources  \\
\small
\textbf{Correspondence:} \texttt{happy.buzaaba@princeton.edu}
}

\begin{document}

\maketitle


\begin{abstract}

Despite their linguistic diversity and global significance, African languages remain underrepresented in research and resources to support NLP. We aim to bridge this gap by introducing \textbf{AfriSUD}, the first multilingual collection of syntactically annotated treebanks for nine diverse African languages spanning major language families and regions across Sub-Saharan Africa. Using the Surface-Syntactic Universal Dependencies (SUD) framework, our community-led effort provides high-quality, native-speaker verified data that capture typological key features such as agglutination and tone. We evaluate a range of models on AfriSUD for part-of-speech tagging and dependency parsing including non-transformer baselines, multilingual pretrained encoders, and LLMs. Our results reveal a significant syntax gap, where models still show clear limitations across the nine languages, suggesting that existing architectures may not fully capture the structural diversity of African-language syntax.
\end{abstract}

\section{Introduction}

African languages account for nearly a third of the world's linguistic diversity, yet they remain disproportionately underrepresented in NLP research. Despite the foundational role of dependency parsing in tasks such as machine translation~\citep{duan2023syntax, li2023neural} and information extraction~\citep{joshi2025dependency, fei2023constructing}, most African languages lack the gold-standard treebanks required for developing and evaluating syntactic models. While large-scale efforts like Universal Dependencies (UD) \cite{nivre-etal-2016-universal} have made significant progress, their coverage of African languages remains limited. Of the 186 languages documented in UD v2.17 \cite{nivre-etal-2020-universal}, less than 15 are African languages or African-based. 
This limited syntactic coverage presents a barrier to building grammar-aware tools, conducting cross-lingual transfer, and performing evaluations of state-of-the-art multilingual models in low-resource settings.

Developing syntactic resources for African languages poses significant challenges, as many languages in the Niger-Congo and Afro-Asiatic families exhibit complex morphological and phonological features, notably agglutination (concatenation of roots and grammatical morphemes), extensive noun class systems, and tonal distinctions affecting part-of-speech membership,  which are poorly supported by standard NLP pipelines and underrepresented in existing annotated corpora~\citep{alabi-etal-2025-charting}. While recent community-driven initiatives have established benchmarks for machine translation \citep{adelani-etal-2022-thousand}, named entity recognition \citep{adelani-etal-2022-masakhaner}, and part-of-speech tagging \citep{dione-etal-2023-masakhapos}, comprehensive syntactic resources for African languages remain largely absent.

In this work, we introduce AfriSUD, the first coordinated, community-driven effort to bridge this gap. We present a collection of syntactically annotated treebanks for nine diverse African languages spanning multiple geographic regions and language families, serving as lingua francas for millions of first- and second-language speakers.
Our work is grounded in the Surface-Syntactic Universal Dependencies (SUD) framework \cite{gerdes-etal-2018-sud, gerdes2021starting}, which we argue is particularly suitable for many African languages. In particular, SUD differs from UD in how it represents grammatical markers, UD generally attaches them to lexical heads, whereas SUD more often represents them as syntactic heads. This distinction is especially relevant for independent Tense–Aspect–Mood (TAM) markers in many Bantu languages, whose structural role can be represented directly in SUD. Importantly, SUD can be deterministically converted to and from UD, ensuring interoperability between the two schemes.
To evaluate AfriSUD as a benchmark for African language syntactic NLP, we compare a diverse set of models including supervised parsers, multilingual pretrained models, Africa-centric encoders and large language models (LLMs) on part-of-speech tagging and dependency parsing across nine African languages. Our results reveal a persistent gap between structure and labeling: models are better at identifying dependency attachments (as between a determiner and a noun head) than dependency relations (as between a verb and an object), especially for language-specific relations.

To summarize, we introduce AfriSUD, a multilingual SUD treebank collection for African languages, annotated and verified by native-language linguists. We also document key annotation challenges under the SUD~\citep{gerdes-etal-2018-sud} framework, including the treatment of clitics and bound morphemes, syntactic and lexical ambiguity, and language-specific relation choices. We benchmark a range of models including supervised parsers, multilingual and Africa-centric encoders, and LLMs, finding a consistent gap between head prediction and relation labeling. Finally, we provide a construction-level analysis showing that labeling errors often concentrate in serial verb constructions, Tense-Aspect-Mood (TAM) auxiliaries, and possessives. We release AfriSUD and annotation guidelines under academic license or CC BY-NC 4.0 on HuggingFace~\footnote{https://huggingface.co/datasets/Lugha-Models/AfriSUD}. Efik is excluded from public release due to restrictions on its source data (see Appendix \ref{sec:appendix_release}).

\section{Related work}

Recent efforts have expanded African-language NLP resources, but syntactic annotation remains comparatively low. Existing dependency treebanks have generally produced individual language resources, as for Bambara \citep{aplonova-tyers-2017-towards}, Amharic \citep{seyoum-etal-2018-universal}, Naija (Nigerian Pidgin)~\citep{caron2019surface}, Wolof \citep{dione2019developing}, Yorùbá \citep{ishola-zeman-2020-yoruba}, Beja~\citep{kahane2021morph}, Khoekhoe~\citep{kira-etal-2025-universal}, and Gbaya~\citep{roulon2025morpheme}. The resulting coverage is limited and spotty, making it difficult to evaluate parsers systematically across African languages. In parallel, community-led initiatives that have produced benchmark datasets for NLP tasks such as named-entity recognition, part-of-speech tagging, and language understanding, include  MasakhaNER~\citep{adelani-etal-2021-masakhaner}, MasakhaPOS~\citep{dione-etal-2023-masakhapos}, IrokoBench~\cite{adelani-etal-2025-irokobench}, and INJONG~\cite{yu-etal-2025-injongo}. Recent Africa-centric models such as AfriBERTa \citep{ogueji-etal-2021-small} and AfroXLMR~\citep{alabi-etal-2022-adapting,adelani-etal-2024-sib}, Lugha-Llama~\citep{buzaaba2025lugha}, and AfriqueLLM~\citep{yu2026afriquellm} further show the value of adapting models to African-language data. AfriSUD complements these efforts by providing a coordinated dependency treebank collection designed for syntactic evaluation across typologically diverse African languages.

\section{AfriSUD Dataset}

\begin{table*}[t]
\centering
\small
\setlength{\tabcolsep}{4pt}
\begin{tabular}{ll l l c c r r r}
\toprule
\textbf{Language} & \textbf{ISO code} & \textbf{Source} 
& \textbf{Genre} 

  & \textbf{Morph.} & \textbf{Tone} 
  & \textbf{Ann. Sents.} & \textbf{Tokens} & \textbf{Avg.\ len} \\
\midrule

Hausa & \texttt{hau} & MasakhaPOS  & News & Fus  & \checkmark & 1158   & 27{,}304 & 23.6 \\

Naija & \texttt{pcm} & MasakhaPOS & News  
& Iso  & -- & 1{,}356 & 33{,}065 & 24.4 \\

Wolof & \texttt{wol} & MasakhaPOS & News & Agg  & -- & 921   & 19{,}438 & 21.1 \\

Yoruba & \texttt{yor} & MasakhaPOS + Menyo-20k & News & Iso & \checkmark & 1028  & 16{,}238 & 15.8 \\

Efik & \texttt{efi} & School Primers & Educational & Agg  & \checkmark & 800   & 8{,}477  & 10.6 \\

Swahili & \texttt{swa} & MasakhaPOS & News & Agg  & -- & 1{,}491 & 37{,}417 & 25.1 \\

Kinyarwanda  & \texttt{kin} & MasakhaPOS & News & Agg  & \checkmark  & 1{,}378 & 33{,}419 & 24.3 \\

Runyankore  & \texttt{nyn} & SALT-MT & Translated & Agg  & -- & 968   & 8{,}479  & 8.8  \\

isiXhosa  & \texttt{xho} & MasakhaPOS  
& News  & Agg  & \checkmark & 464   & 7{,}931  & 17.1 \\
\bottomrule
\end{tabular}
\caption{\textbf{AfriSUD language and corpus statistics.} We report corpus source, genre, morphological type (Iso = isolating, Agg = agglutinative, Fus = fusional), tone, and treebank statistics after quality assurance, including annotated sentences (Ann.\ Sents.), and tokens. Tokens refer to orthographic word tokens used for syntactic annotation. A full CoNLL-U/SUD
annotation example for Wolof is provided in Appendix Table~\ref{tab:conllu-example-wol}}
\label{tab:language_stats}
\end{table*}
The AfriSUD dataset contains treebanks for nine African languages spanning West, East, and Southern Africa and representing three linguistic families: Niger-Congo, Afroasiatic and English-based creoles. Within Niger–Congo, four languages (Swahili, Kinyarwanda, Runyankore, and IsiXhosa) belong to the Bantu subgroup, Wolof is a member of the Senegambian subgroup, while Yorùbá and Efik belong to the Volta–Niger and Cross River subgroups, respectively ~\citep{nurse2003towards}.
Hausa belongs to the Chadic branch of the Afroasiatic family, while Naija is an English-based creole.
This geographic, genealogical, and typological coverage reflects the linguistic diversity of the African continent and is in line with recent efforts to build multilingual benchmarks for African languages \citep{adelani-etal-2021-masakhaner, dione-etal-2023-masakhapos, adelani-etal-2025-irokobench}. 

The languages in the dataset differ substantially with regard to morphosyntax. Most of them are agglutinative, particularly the Bantu languages, where complex verbal morphology can encode multiple grammatical categories within a single word. In contrast, languages such as Yorùbá and Naija are isolating, relying primarily on word order and particles to express grammatical relations, while Hausa exhibits fusional morphology, where roots and morphemes are not clearly separated as in agglutinative languages. These differences affect how syntactic information is distributed across tokens and pose known challenges for dependency parsing and multilingual modeling \citep{ogueji-etal-2021-small, adebara-abdul-mageed-2022-towards}.
Hausa, Yorùbá, Efik, Kinyarwanda, and IsiXhosa use lexical tones to differentiate meanings on an otherwise identical form, and this may affect the part-of-speech assignment. Bantu languages distinguish between ten and twenty different noun classes, which govern agreement between heads and dependents \citep{katamba2003bantu, babou2016noun} and interact with verbal and nominal morphology \citep{dione-etal-2023-masakhapos}, affecting syntactic structure. Together, these properties make the AfriSUD languages a useful testbed for evaluating syntactic models across typologically diverse African languages.

Table~\ref{tab:language_stats} reports AfriSUD treebank statistics after quality assurance, including the source and genre of the corpus. All corpora consist of written text. Seven languages contain data from MasakhaPOS~\citep{dione-etal-2023-masakhapos}, a widely used African-language benchmark covering news and general-domain text. Yorùbá includes additional data from  Menyo-20k~\citep{adelani-etal-2021-effect}, a general-domain machine translation corpus. The Efik corpus comes from written school primers, while the Runyankore corpus comes from SALT-MT translated text~\citep{akera2022machine}.
Beyond corpus size, treebanks also vary in their structural properties. The number of distinct dependency relations ranges from 29 in Wolof to 69 in IsiXhosa, while the proportion of non-projective arcs ranges from 0.01\% in Hausa to 3.63\% in IsiXhosa. SUD-distinctive constructions are highly language-specific: \texttt{comp:poss} occurs mainly in Kinyarwanda, while \texttt{comp:cleft} is largely restricted to IsiXhosa. The underspecified \texttt{udep} relation ranges from 0.0 to 186.9 occurrences per 1,000 tokens. More statistics on relation distributions, rare labels, tree depth, dependency length, non-projectivity, and construction frequencies are reported in the Appendix~\ref{sec:Structural_profile}.

\subsection{Annotation Methodology}

\definecolor{posAUX}{RGB}{253,192,134}    
\definecolor{posVERB}{RGB}{178,223,138}   
\definecolor{posDET}{RGB}{202,178,214}    
\definecolor{posNOUN}{RGB}{166,206,227}   
\definecolor{posCCONJ}{RGB}{255,255,153}  
\definecolor{posADP}{RGB}{204,235,197}    
\definecolor{posPUNCT}{RGB}{220,220,220}  
\definecolor{posPRON}{RGB}{253,224,239}   
\definecolor{posADV}{RGB}{255,217,102}    
\definecolor{posPROPN}{RGB}{188,189,220}  
\definecolor{posSCONJ}{RGB}{251,180,174}  
\definecolor{posPART}{RGB}{229,245,224}   
\definecolor{glossColor}{RGB}{153,0,0} 

\newcommand{\pos}[2]{%
  \tikz[baseline=(n.base)]{%
    \node[draw, rounded corners=2pt, fill=#1,
          inner xsep=3pt, inner ysep=1.5pt,
          font=\fontsize{6.5}{7}\selectfont\bfseries] (n) {#2};}}

\begin{figure*}[t]
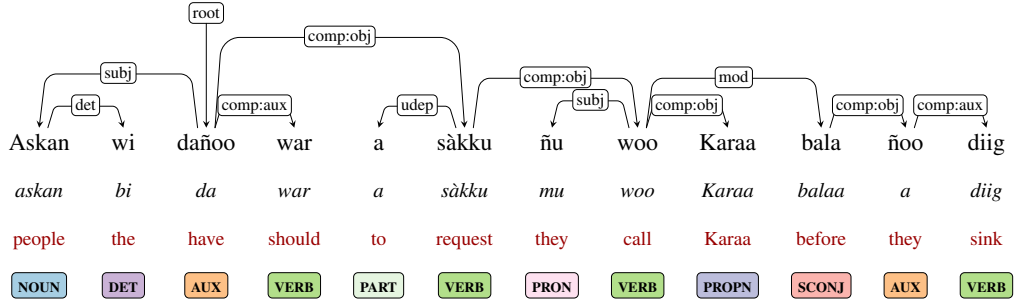

\centering
\scalebox{0.82}{%
\begin{dependency}
  \begin{deptext}[column sep=0.9em, row sep=0.75em]
    Askan \& wi \& dañoo \& war \& a \& sàkku \& ñu \& woo \&
    Karaa \& bala \& ñoo \& diig 
    \\
    {\fontsize{9.5}{9}\selectfont\textit{askan}}  \&
    {\fontsize{9.5}{9}\selectfont\textit{bi}}     \&
    {\fontsize{9.5}{9}\selectfont\textit{da}}     \&
    {\fontsize{9.5}{9}\selectfont\textit{war}}    \&
    {\fontsize{9.5}{9}\selectfont\textit{a}}      \&
    {\fontsize{9.5}{9}\selectfont\textit{sàkku}}  \&
    {\fontsize{9.5}{9}\selectfont\textit{mu}}     \&
    {\fontsize{9.5}{9}\selectfont\textit{woo}}    \&
    {\fontsize{9.5}{9}\selectfont\textit{Karaa}}  \&
    {\fontsize{9.5}{9}\selectfont\textit{balaa}}  \&
    {\fontsize{9.5}{9}\selectfont\textit{a}}      \&
    {\fontsize{9.5}{9}\selectfont\textit{diig}}   \&
    \\
    {\fontsize{9}{9}\selectfont\textcolor{glossColor}{people}}      \&
    {\fontsize{9}{9}\selectfont\textcolor{glossColor}{the}}         \&
    {\fontsize{9}{9}\selectfont\textcolor{glossColor}{have}}        \&
    {\fontsize{9}{9}\selectfont\textcolor{glossColor}{should}}      \&
    {\fontsize{9}{9}\selectfont\textcolor{glossColor}{to}}          \&
    {\fontsize{9}{9}\selectfont\textcolor{glossColor}{request}}     \&
    {\fontsize{9}{9}\selectfont\textcolor{glossColor}{they}}        \&
    {\fontsize{9}{9}\selectfont\textcolor{glossColor}{call}}        \&
    {\fontsize{9}{9}\selectfont\textcolor{glossColor}{Karaa}}       \&
    {\fontsize{9}{9}\selectfont\textcolor{glossColor}{before}}      \&
    {\fontsize{9}{9}\selectfont\textcolor{glossColor}{they}}        \&
    {\fontsize{9}{9}\selectfont\textcolor{glossColor}{sink}}       
    \\
    \pos{posNOUN}{NOUN}   \& \pos{posDET}{DET}     \& \pos{posAUX}{AUX}    \&
    \pos{posVERB}{VERB}   \& \pos{posPART}{PART}   \& \pos{posVERB}{VERB}  \&
    \pos{posPRON}{PRON}   \& \pos{posVERB}{VERB}   \& \pos{posPROPN}{PROPN}\&
    \pos{posSCONJ}{SCONJ} \& \pos{posAUX}{AUX}     \& \pos{posVERB}{VERB}  
    \\
  \end{deptext}

  \deproot[edge unit distance=3.0ex]{3}{root}
  \depedge[edge unit distance=2.5ex]{3}{1}{subj}
  \depedge[edge unit distance=2.0ex]{1}{2}{det}
  \depedge[edge unit distance=2.0ex]{3}{4}{comp:aux}
  \depedge[edge unit distance=2.8ex]{3}{6}{comp:obj}
  \depedge[edge unit distance=2.0ex]{6}{5}{udep}
  \depedge[edge unit distance=2.3ex]{6}{8}{comp:obj}
  \depedge[edge unit distance=2.5ex]{8}{7}{subj}
  \depedge[edge unit distance=2.0ex]{8}{9}{comp:obj}
  \depedge[edge unit distance=2.3ex]{8}{10}{mod}
  \depedge[edge unit distance=2.0ex]{10}{11}{comp:obj}
  \depedge[edge unit distance=2.0ex]{11}{12}{comp:aux}

\end{dependency}}

\caption{\textbf{Example annotation of a Wolof sentence in AfriSUD}
  (\textit{``The people must demand that Karaa be called in before the team goes under.''}),
  each token shows the word form, lemma in italics, morpheme gloss in red, and POS tags in a coloured box. The arcs indicate dependency heads and relation labels.}
\label{fig:wol-deptree}
\end{figure*}
AfriSUD annotation follows the Surface-Syntactic Universal Dependencies (SUD) framework \citep{gerdes-etal-2018-sud}, which represents syntactic relations close to surface structure. This proves useful for the target languages where auxiliaries and other functional elements often encode tense, aspect, mood, and agreement information. The annotation pipeline includes lemmatization, Universal part-of-speech (UPOS) tagging, dependency head annotation, and dependency relation labeling, as illustrated with a Wolof example in Figure~\ref{fig:wol-deptree}. We use the 17 standard UPOS tags \citep{petrov-etal-2012-universal} and an SUD relation set covering core relations such as subjects, complements, and modifiers, as well as constructions such as auxiliary complements (\texttt{comp:aux}), predicative complements (\texttt{comp:pred}), and serial verb constructions (\texttt{compound:svc}). Appendix Table~\ref{tab:annotation_inventory} provides the full set of POS tags and dependency relations used in our annotation, along with their definitions.

Annotation was conducted using ArboratorGrew~\citep{guibon-etal-2020-collaborative} and AfriSUD languages with existing treebanks for Naija, Wolof, and Yorùbá. We used the integrated parser interface in ArboratorGrew\footnote{https://arborator.grew.fr}, based on BertForDeprel~\citep{guiller2020bertfordeprel} to generate initial pre-annotations from the available data. The pre-annotations were reviewed and manually corrected by annotators, while the remaining languages were annotated from scratch. Across all languages, annotation follows SUD conventions: auxiliaries are treated as syntactic heads, noun class information is encoded as UFeats, serial verb constructions are annotated with \texttt{compound:svc}, and underspecified relations are marked with \texttt{udep} when necessary.

\subsection{Quality Control}

Each language was annotated by three native-speakers: a coordinator and two annotators. All annotators completed training sessions on dependency grammar, the SUD framework, language-specific guidelines, and a pilot annotation exercise of ten sentences per language. Since dependency annotation requires interdependent decisions about tokenization, heads, and relation labels, standard Inter-Annotator Agreement metrics such as Fleiss's Kappa could not be straightforwardly computed or interpreted. Following prior African-language annotation work \citep{dione-etal-2023-masakhapos} we adopt a consensus-based adjudication procedure to resolve disagreements. Nonetheless, in Table~\ref{tab:iaa} we report inter-annotator agreement for Efik, Runyankore, Kinyarwanda, and Yorùbá for which a sufficiently large set of sentences was independently annotated by both annotators. Overlap was too limited for reliable estimates in the remaining languages. As shown in Table~\ref{tab:iaa}, agreement is high across the four languages, with $\kappa_{\text{UPOS}}$ ranging from 0.88 in Kinyarwanda to 0.96 in Runyankore, consistent with the verb-morphology ambiguity discussed in \S\ref{sec:syntaxlexambig}. LAS is consistently lower than UAS, consistent with the gap between head attachment and relation labeling observed in model predictions (\S\ref{sec:results},\S\ref{sec:error-analysis}) and suggesting that relation labeling is more challenging than identifying the correct head.

\begin{table}[t]
\centering
\resizebox{\columnwidth}{!}{%
\begin{tabular}{@{}lrrrrrrr@{}}
\toprule
\textbf{Lang.} & \textbf{Sents} & \textbf{Tok.} & \textbf{UPOS} & \boldmath{$\kappa_{\text{U}}$} & \textbf{UAS} & \textbf{LAS} & \boldmath{$\kappa_{\text{d}}$} \\
\midrule
Efik        & 1,300 & 19,784 & 95.2 & 0.94 & 88.6 & 86.1 & 0.87 \\
Runyankore  & 975   & 8,654  & 97.0 & 0.96 & 95.9 & 94.7 & 0.95 \\
Kinyarwanda & 384   & 8,977  & 89.6 & 0.88 & 82.2 & 76.0 & 0.77 \\
Yor\`ub\'a  & 110   & 1,815  & 82.0 & 0.79 & 89.6 & 79.8 & 0.80 \\
\bottomrule
\end{tabular}%
}
\caption{Inter-annotator agreement for languages with sufficient annotation overlap. Sents and Tok. denote the number of sentences and tokens independently annotated by both annotators. UPOS is raw agreement, $\kappa_{\text{U}}$ Cohen’s kappa for UPOS, UAS/LAS attachment agreement, and $\kappa_{\text{d}}$ Cohen’s kappa for dependency labels.}
\label{tab:iaa}
\end{table}
The language coordinators supervised the annotation and held regular discussions with the annotators to resolve ambiguous cases and disagreements so as to ensure consistency with the SUD guidelines across languages. After adjudication, each sentence received a single final annotation agreed upon by the language team. We then applied automatic validation checks to detect malformed dependency structures, including missing part-of-speech or Dependency relation values, absent or multiple roots, root-label/head mismatches, and cycles in the dependency graph. Annotators and coordinators were compensated for their work.\footnote{Each annotator was paid US\$750.}

\subsection{Annotation Challenges}

This section analyzes the annotation challenges encountered when applying the SUD formalism to our African languages. The analysis is structured around three major issues: (i) clitics and morphological binding, (ii) ambiguity resolution, (iii) language-specific challenges and SUD relations. 

\subsubsection{Clitics and morphological binding}
A central difficulty arises from the  rich morphological structure where grammatical information is encoded in  affixes or clitic-like elements that are tightly bound to lexical stems. Agglutination is repeatedly identified as a critical issue in most of the languages. Two distinct methodologies emerged from the data: morphological decomposition (considering roots and the attached morphemes separately) vs.\  single-token preservation.
The Efik, Yorùbá, and Kinyarwanda teams opted to segment orthographic words into syntactic components. Efik, for instance, is an agglutinating language, reflecting a one-to-one correspondence of morpheme to meaning, with these morphemes often arranged in a specific linear order. For example, the verb \textit{emetem} (``You had cooked it'') is decomposed into the second-person pronoun \textit{e-}, the past tense marker \textit{me-}, and the verb root \textit{tem}. Likewise, in Yoruba, fused prepositional constructions such as `\textit{sílé}' and `\textit{níta}' were restored to their base forms `\textit{sí} \textit{ilé}' (in house) and `\textit{ní} \textit{ìta}' (in outside).

Conversely, IsiXhosa, Swahili, and Hausa maintained lexical integrity. In IsiXhosa, complex verbs containing subject markers (\textit{ndi-}, \textit{u-}) and tense markers (\textit{ya-}) are treated as single tokens. The Hausa team treated derived nouns, such as \textit{ma'aurata},``married couple'' (from \textit{aure}, ``marry''), as a single unit rather than splitting off the nominalizing prefix \textit{ma-}. This decision aligns with the surface-oriented approach of SUD. In practice, morphological information is encoded through features rather than syntactic dependencies, thereby preserving structural consistency while avoiding over-segmentation.

\subsubsection{Syntactic and Lexical Ambiguities}

\label{sec:syntaxlexambig}
Ambiguity is a pervasive issue across the dataset on multiple levels, including lexical, morphological, and syntactic ambiguity. 
The dominant strategy for ambiguity resolution relied heavily on context-driven analysis and established team-level annotation conventions. Annotators consistently favor interpretations that are semantically plausible and structurally coherent.

For instance, in Yorùbá, the morpheme \textit{n\'i} is multifunctional (copula, main verb, or conjunction). If \textit{n\'i} follows a subject, it is annotated as a verb; if it heads a subordinate clause, it is treated as a subordinating conjunction. In Kinyarwanda, ambiguity is especially prominent in verb morphology, where a single form may encode multiple grammatical functions. Words like \textit{gukora} can be a verb (``to do'') or a noun (``the act of doing''), requiring the examination of surrounding tense markers to determine the correct part of speech.
In Efik, ambiguity arises in complex sentence structures involving modifier clauses, cleft sentences, and multiple embedded subjects and objects. Having recognized the relevant patterns, annotation often required identifying the presence of similar features in each successive sentence and annotating accordingly. The annotators here relied on shared guidelines to ensure consistency and a uniform annotation pattern across the corpus.

\subsubsection{Language-Specific Challenges and SUD Relations}

\label{sec:langspecific}
Beyond shared challenges, each language presents unique difficulties that impact the annotation process. In addition, several SUD relations prove difficult to apply consistently across languages, particularly in contexts involving complex syntax or rich morphology. 

Some of the problems can be described as follows. 
Head (or root) selection in a language like Yorùbá is not always easy as the language employs multiple auxiliaries (e.g., aspectual markers \textit{ti} and \textit{\`n}, or future markers \textit{y\'o\`o}). \footnote{It should be noted that SUD is based on distributional criteria and treats function words as heads, unlike UD. For instance, adpositions (ADP) are heads of adpositional phrases, auxiliaries (AUX) are heads of complex verbal forms, and subordinating conjunctions (SCONJ) are heads of subordinated clauses.} The convention among the annotators was to establish a hierarchy for the auxiliaries and to select the one that precedes the others as the rightful head. In an isolating language like Yorùbá where pluralization is not derived through inflection, the marker \textit{\`aw\d{o}n} is added before a noun to mark plural. Annotating this syntactic relationship in SUD was not straightforward. Annotators decided to use the relationship ``compound:prt" which seems to be closely related to the syntactic relationship found between nouns and the plural marker. 
Furthermore, problems arose with the underspecified \textit{udep} relation, which covers both comp and mod in cases where a dependent cannot be clearly classified as an argument or a modifier, or be specified as comp or mod.  
Some languages show copula constructions without an overt copula verb like English "to be". For instance, in IsiXhosa, the absence of copula constructions led to the treatment of the predicate as the head and the marker as a copula. 

In summary, while the SUD framework provides a useful baseline for cross-linguistic annotation, its application to morphologically rich and under-resourced African languages requires careful adaptation. 
Common challenges include the treatment of bound morphemes, pervasive ambiguity, and structural mismatches between linguistic phenomena and formal annotation schemes. Addressing these challenges requires a combination of theoretical flexibility, empirical observation, and collaborative annotation practices. Annotators must adopt a combination of normalization procedures and language-specific adaptations of SUD guidelines. Iterative annotation and validation cycles were frequently used to refine decisions. 
In some cases, annotators explicitly acknowledged the need to deviate slightly from standard guidelines to better reflect the linguistic reality of the language.

\section{Experiments Setup}
\subsection{Baseline Models}

\paragraph{Stanza}
We use Stanza~\citep{qi-etal-2020-stanza}, a neural dependency parsing pipeline, as a strong non-transformer baseline. We initialize the parser with pretrained fastText embeddings~\citep{grave-etal-2018-learning} when available. For Runyankore (\texttt{nyn}) and Efik(\texttt{efi}), which are not covered by the pretrained embeddings, we train new embeddings from publicly available corpora using SALT~\citep{akera2022machine} for \texttt{nyn} and MT560~\citep{gowda-etal-2021-many} together with SIB-200 Ibom~\citep{kalejaiye-etal-2025-ibom} for \texttt{efi}.
\paragraph{Multilingual encoders}
For the transformer baselines, we fine-tune an end-to-end biaffine dependency parser~\citep{dozat2016deep} with pretrained encoders: (1) general multilingual models mBERT~\citep{devlin-etal-2019-bert} and XLM-RoBERTa
Large~\citep{conneau-etal-2020-unsupervised} and (2) Africa-centric models AfriBERTa-large~\citep{ogueji-etal-2021-small}, 
AfroXLMR-large~\citep{alabi-etal-2022-adapting}, and AfroXLMR-large-76L~\citep{adelani-etal-2024-sib}. In the experiments, each treebank is split into 70/10/20 for train/dev/test partitions. We report unlabeled attachment score (UAS) and labeled attachment score (LAS), which measure correct head assignment and correct head-plus-label prediction, respectively. All transformer-based models are fine-tuned with HuggingFace Transformers~\citep{wolf-etal-2020-transformers}, using a maximum sequence length of 512, batch size 16, gradient accumulation 2, learning rate $5\times10^{-5}$, and 50 epochs on a single A100 NVIDIA GPU.

\subsection{LLM Prompting}

We evaluate widely used LLMs: Gemini-3.1-Pro~\citep{gemini31pro2026}, GPT-5.2\footnote{https://developers.openai.com/api/docs/models/gpt-5.2}, GPT-4o~\citep{hurst2024gpt}, and Gemma-3-12B-IT/27B-IT~\citep{gemmateam2025gemma3technicalreport}. All models are evaluated with deterministic decoding by setting the temperature to $\tau=0$. For open-weight models, we use greedy decoding and set the maximum number of generated tokens to 2,048. In our experiments, the task is formulated as structured generation: given the raw sentence text and pre-segmented tokens (id and surface form), the model predicts each token's lemma, UPOS tag, syntactic head, and dependency relation. A single prompt template is used across all models and languages to ensure fair comparison. The complete prompt and output schema are provided in Figure~\ref{fig:llm_prompt_template}.
We perform zero-shot and few-shot prompting with $K \in \{0,1,5\}$ demonstrations. Few-shot examples are sampled from a held-out pool comprising 10\% of each language's data, which is reserved exclusively for demonstrations and excluded from evaluation. For both 1-shot and 5-shot settings, we use five different demonstration sets sampled with seeds 13--17 and report the mean and standard deviation across runs.

In addition, we performed supervised fine-tuning using gemma-3-12B for 5 epochs using a learning rate of $1\times10^{-5}$. The SFT dataset was obtained by aggregating training samples from all nine AfriSUD languages. Each example is formatted as a prompt--completion pair, mapping a raw sentence to its gold CoNLL-U/SUD annotation. We use a fixed random seed (42) to assign an instruction template to each training sentence. All instruction templates used are provided in Appendix~\ref{tab:sft_templates_full}.

\subsection{Cross-lingual Transfer}

Cross-lingual transfer depends on several factors, including model choice, transfer strategy, and selection of an appropriate source language. Previous work on cross-lingual dependency parsing shows that in the zero-shot setting, the choice of the source language is important, especially when the source and target languages are typologically distant~\citep{tran-bisazza-2019-zero,agic-2017-cross}. Although English is commonly used as the transfer source due to resource availability, evidence from cross-lingual syntactic transfer indicates that better transfer can often be obtained from sources that are structurally closer to the target language~\citep{duong-etal-2015-cross}. In addition, studies in cross-lingual syntax indicate that transfer performance depends in part on the relationship between source and target languages, including their typological similarity and broader structural proximity~\citep{litschko-etal-2020-towards, fisch-etal-2019-working}.

We consider seven source languages for cross-lingual syntactic transfer: English (\texttt{eng}), French (\texttt{fra}), Afrikaans (\texttt{afr}), Arabic (\texttt{ara}), Romanian (\texttt{ron}), Naija (\texttt{pcm}), and Wolof (\texttt{wol}). These source languages were selected based on supervised SUD treebank availability and typological diversity, including variation in word order and the relative order of syntactic heads and their dependents, which are known to affect cross-lingual dependency parsing~\citep{scholivet-etal-2019-typological,liu-etal-2020-cross-lingual-dependency}. 
For Wolof and Naija, we further evaluate augmented variants, denoted \texttt{+wtb} and \texttt{+nsc}, respectively. In these settings, the source-language training data are supplemented with pre-existing SUD treebanks from the SUD release~\footnote{https://surfacesyntacticud.org/data/}:\texttt{\detokenize{SUD_Wolof-WTB@2.17}} and \texttt{\detokenize{SUD_Naija-NSC@2.17}}.
Romanian is included in light of previous work showing that the choice of source language can substantially affect cross-lingual transfer performance in dependency parsing \citep{agic-2017-cross, dione-etal-2023-masakhapos}.

\section{Results}
\label{sec:results}
\subsection{Baseline results}
\begin{table*}[t]
\centering
\scriptsize
\setlength{\tabcolsep}{4pt}
\renewcommand{\arraystretch}{1.0}
\begin{tabular}{l|l|rrrrrrrrr |c}
\toprule
\textbf{Metric} & \textbf{Model} & \textbf{efi} & \textbf{hau} & \textbf{kin} & \textbf{nyn} & \textbf{pcm} & \textbf{swa} & \textbf{wol} & \textbf{xho} & \textbf{yor} & \textbf{Avg} \\
\midrule
\multirow{6}{*}{\textbf{UPOS Acc.}}
& Stanza                       & 83.3 & 91.6 & 95.8 & 84.4 & 81.4 & 91.3 & 91.8 & 84.5 & 91.3 & 88.4 {\scriptsize $\pm$ 0.5} \\
& mBERT                        & 81.8 & 90.1 & 95.2 & 83.6 & 90.5 & 90.9 & 90.3 & 82.1 & 91.3 & 88.4 {\scriptsize $\pm$ 0.4} \\
& XLM-RoBERTa$_\text{large}$         & 82.8 & 92.4 & 95.8 & 84.4 & 90.8 & 91.8 & 91.2 & 84.8 & 91.5 & 89.5 {\scriptsize $\pm$ 0.3} \\
& AfriBERTa$_\text{large}$     & 82.1 & \textbf{92.7} & 96.0 & 85.0 & 90.2 & 92.1 & 90.0 & 81.4 & 91.5 & 89.0 {\scriptsize $\pm$ 0.3} \\
& AfroXLM-R$_\text{large}$     & 84.5 & 92.6 & \textbf{96.4} & 75.9 & \textbf{91.1} & \textbf{92.5} & 91.6 & 86.6 & 91.6 & 89.2 {\scriptsize $\pm$ 0.3} \\
& AfroXLM-R$_\text{large-76L}$ & \textbf{84.9} & \textbf{92.7} & 96.3 & \textbf{86.8} & 91.0 & 92.3 & \textbf{92.6} & \textbf{86.8} & \textbf{91.8} & \textbf{90.6} {\scriptsize $\pm$ 0.2} \\
\cmidrule(lr){1-12}
\multirow{6}{*}{\textbf{UAS}}
& Stanza                       & \textbf{68.8} & 92.8 & \textbf{86.6} & \textbf{86.7} & \textbf{82.5} & \textbf{89.7} & 85.2 & \textbf{82.6} & 85.0 & \textbf{84.4} {\scriptsize $\pm$ 0.3} \\
& mBERT                        & 61.8 & 89.2 & 84.2 & 82.1 & 80.1 & 84.9 & 79.3 & 65.1 & 83.4 & 78.9 {\scriptsize $\pm$ 0.3} \\
& XLM-RoBERTa$_\text{large}$         & 63.3 & 92.6 & 85.1 & 83.3 & 81.5 & 87.2 & 81.2 & 74.6 & 82.7 & 81.3 {\scriptsize $\pm$ 0.5} \\
& AfriBERTa$_\text{large}$     & 58.1 & 91.7 & 84.7 & 82.9 & 79.6 & 86.2 & 76.0 & 63.2 & 84.8 & 78.6 {\scriptsize $\pm$ 0.5} \\
& AfroXLM-R$_\text{large}$     & 66.2 & \textbf{93.1} & 85.7 & 86.4 & 81.3 & 87.7 & 81.9 & 74.1 & 85.9 & 82.5 {\scriptsize $\pm$ 0.4} \\
& AfroXLM-R$_\text{large-76L}$ & 66.5 & \textbf{93.1} & 86.0 & \textbf{86.7} & 81.7 & 87.5 & \textbf{86.2} & 73.8 & \textbf{86.2} & 83.1 {\scriptsize $\pm$ 0.3} \\
\cmidrule(lr){1-12}
\multirow{6}{*}{\textbf{LAS}}
& Stanza                       & \textbf{58.3} & 81.3 & \textbf{79.0} & \textbf{80.7} & \textbf{77.3} & \textbf{84.6} & \textbf{80.0} & \textbf{77.1} & 79.2 & \textbf{77.5} {\scriptsize $\pm$ 0.5} \\
& mBERT                        & 48.9 & 75.3 & 73.8 & 71.3 & 71.5 & 73.9 & 72.1 & 56.2 & 76.1 & 68.8 {\scriptsize $\pm$ 0.4} \\
& XLM-RoBERTa$_\text{large}$         & 50.5 & 80.9 & 75.1 & 72.7 & 73.1 & 76.7 & 74.3 & 65.2 & 75.2 & 71.5 {\scriptsize $\pm$ 0.6} \\
& AfriBERTa$_\text{large}$     & 46.1 & 80.4 & 75.3 & 72.9 & 70.9 & 75.6 & 68.8 & 54.3 & 77.4 & 69.1 {\scriptsize $\pm$ 0.5} \\
& AfroXLM-R$_\text{large}$     & 53.5 & \textbf{81.8} & 76.6 & 76.8 & 73.0 & 77.4 & 75.1 & 66.5 & 79.0 & 73.3 {\scriptsize $\pm$ 0.5} \\
& AfroXLM-R$_\text{large-76L}$ & 53.9 & \textbf{81.8} & 76.8 & 77.4 & 73.6 & 77.1 & 79.6 & 65.9 & \textbf{79.3} & 73.9 {\scriptsize $\pm$ 0.34} \\
\bottomrule
\end{tabular}
\caption{\textbf{UPOS tagging and dependency parsing performance on AfriSUD}. Scores are averaged over five runs; Avg reports the macro-average across languages with mean per-language standard deviation. Bold indicates the best result for each language and metric.}
\label{tab:combined_results}
\end{table*}

Table~\ref{tab:combined_results} shows that transformer-based encoders improve part-of-speech tagging, with AfroXLMR-${\text{large-76L}}$ achieving the best macro-average accuracy of 90.6 compared to 88.4 for Stanza. However, for dependency parsing, Stanza remains a competitive baseline with the highest average UAS score of 84.4 and the highest average LAS of 77.5. 
Among encoder-based models, Africa-centric models outperform general multilingual encoders overall with AfroXLMR-${\text{large-76L}}$ reaching 83.1 UAS and 73.9 LAS while also achieving the top LAS scores for several languages, including Hausa and Yorùbá.

Across all models, LAS is consistently lower than UAS, indicating that relation labeling remains more difficult than identifying dependency heads. Overall, Africa-centric encoders provide the best transformer-based results, particularly for part-of-speech tagging, but Stanza still achieves the best average dependency parsing performance, including the highest LAS, possibly reflecting the stability of parser-specific architectures for relation labeling in low-resource settings.

\subsection{LLMs Prompting results}
\label{sec:prompting-results}

\begin{table}[t]
\centering
\scriptsize
\setlength{\tabcolsep}{2.5pt}
\renewcommand{\arraystretch}{1.0}
\begin{tabular}{llrrrr}
\toprule
\textbf{Model} & \textbf{\# shot} & \textbf{UPOS Acc} & \textbf{UAS} & \textbf{LAS} \\
\midrule
\multirow{3}{*}{Gemini-3.1-Pro}
& 0 & 80.2 \scriptsize{$\pm$0.4} & 65.8 \scriptsize{$\pm$0.3} & 42.4 \scriptsize{$\pm$0.7} \\
& 1 & 82.6 \scriptsize{$\pm$0.2} & 69.7 \scriptsize{$\pm$1.0} & 51.9 \scriptsize{$\pm$0.4} \\
& 5 & \textbf{86.7} \scriptsize{$\pm$0.3} & \textbf{73.0} \scriptsize{$\pm$1.0} & \textbf{59.2} \scriptsize{$\pm$0.8} \\
\midrule
\multirow{3}{*}{GPT-5.2}
& 0 & 76.4 \scriptsize{$\pm$0.4} & 40.4 \scriptsize{$\pm$0.5} & 16.5 \scriptsize{$\pm$0.7} \\
& 1 & 79.5 \scriptsize{$\pm$0.6} & 56.5 \scriptsize{$\pm$1.3} & 39.9 \scriptsize{$\pm$1.4} \\
& 5 & 83.8 \scriptsize{$\pm$0.7} & 66.5 \scriptsize{$\pm$0.9} & 52.5 \scriptsize{$\pm$1.6} \\
\midrule
\multirow{3}{*}{GPT-4o}
& 0 & 72.3 \scriptsize{$\pm$0.4} & 36.1 \scriptsize{$\pm$0.3} & 8.0 \scriptsize{$\pm$0.1} \\
& 1 & 75.2 \scriptsize{$\pm$0.3} & 46.1 \scriptsize{$\pm$0.5} & 28.9 \scriptsize{$\pm$0.5} \\
& 5 & 78.6 \scriptsize{$\pm$0.1} & 52.8 \scriptsize{$\pm$0.2} & 37.3 \scriptsize{$\pm$0.3} \\
\midrule
\multirow{3}{*}{Gemma-3-12B}
& 0 & 60.1 \scriptsize{$\pm$4.3} & 31.2 \scriptsize{$\pm$1.7} &  2.9 \scriptsize{$\pm$0.1} \\
& 1 & 64.3 \scriptsize{$\pm$2.9} & 41.9 \scriptsize{$\pm$2.1} & 17.0 \scriptsize{$\pm$1.2} \\
& 5 & 66.6 \scriptsize{$\pm$3.5} & 48.2 \scriptsize{$\pm$2.4} & 33.9 \scriptsize{$\pm$2.1} \\
\midrule
\multirow{3}{*}{Gemma-3-27B}
& 0 & 64.4 \scriptsize{$\pm$1.6} & 38.5 \scriptsize{$\pm$1.5} &  6.1 \scriptsize{$\pm$0.2} \\
& 1 & 67.9 \scriptsize{$\pm$2.1} & 45.0 \scriptsize{$\pm$2.6} & 16.2 \scriptsize{$\pm$1.1} \\
& 5 & 72.7 \scriptsize{$\pm$2.1} & 51.8 \scriptsize{$\pm$2.8} & 35.1 \scriptsize{$\pm$2.4} \\
\midrule
\midrule
\multicolumn{2}{l}{\textit{supervised baseline}} \\
Gemma-3-12B$_\textsc{sft}$ & SFT & 80.9 \scriptsize{$\pm$3.4} & 66.9 \scriptsize{$\pm$7.1} & 58.0 \scriptsize{$\pm$6.4} \\
Stanza & NIL & 88.4 \scriptsize{$\pm$0.5} & 84.4 \scriptsize{$\pm$0.3} & 77.5 \scriptsize{$\pm$0.5} \\
\bottomrule
\end{tabular}
\caption{\textbf{Average POS accuracy, UAS, and LAS across AfriSUD languages.} Scores are percentages averaged over five runs, the best result for each metric is shown in bold.}
\label{tab:avg_llm_results}
\end{table}
Table~\ref{tab:avg_llm_results} reports the average scores for UPOS, UAS, and LAS in the nine languages. In-context demonstrations consistently improve performance across all models and metrics, with the largest gains observed for LAS. The gains are particularly clear for models with lower zero-shot LAS, i.e., GPT-5.2 improves from 16.5 to 52.5 LAS, while Gemma-3-12B rises from 2.9 to 33.9. The gains extend beyond LAS, additional demonstrations also improve UAS and UPOS accuracy across models. The results show a performance gap between closed and open models. Under 5-shot setting, Gemini-3.1-Pro performs best overall, reaching 86.7 UPOS, 73.0 UAS, and 59.2 LAS followed by GPT-5.2, while GPT-4o and the open-weight Gemma models perform lower particularly on LAS. Across models, the consistent UAS--LAS gap indicates that models identify syntactic heads more reliably than they assign SUD relation labels.

A similar trend is observed in the language-level breakdown, Gemini-3.1-Pro achieves the best 5-shot performance across all languages, with Wolof, Nyankore, Swahili, and Yoruba among the best-performing languages while Efik and Xhosa remain the most challenging. Overall, part-of-speech  tagging achieves higher scores than dependency parsing, while SUD relation labeling remains more challenging. Detailed results per-language are provided in the Appendix Table~\ref{tab:pos_uas_las_llm}.
\paragraph{Few-shot prompting vs. supervised fine-tuning (SFT) for dependency parsing:} Although LLMs generally improve with additional shots, a large gap remains between few-shot prompting and SFT. Compared with a 5-shot Gemma-3-12B, SFT improves performance by $+14.3$ UPOS, $+18.7$ UAS, $+24.1$ LAS. Despite being based on a much smaller model, SFT Gemma-3-12B reduces the gap to 5-shot Gemini-3.1-Pro, with a remaining difference of $1.2$ LAS points. We report the complete results per-language in Table~\ref{tab:pos_uas_las_llm} of the Appendix.

\subsection{Cross-lingual Transfer results}
 
 \begin{figure*}[t]
   \centering
   \includegraphics[width=0.9\textwidth]
   {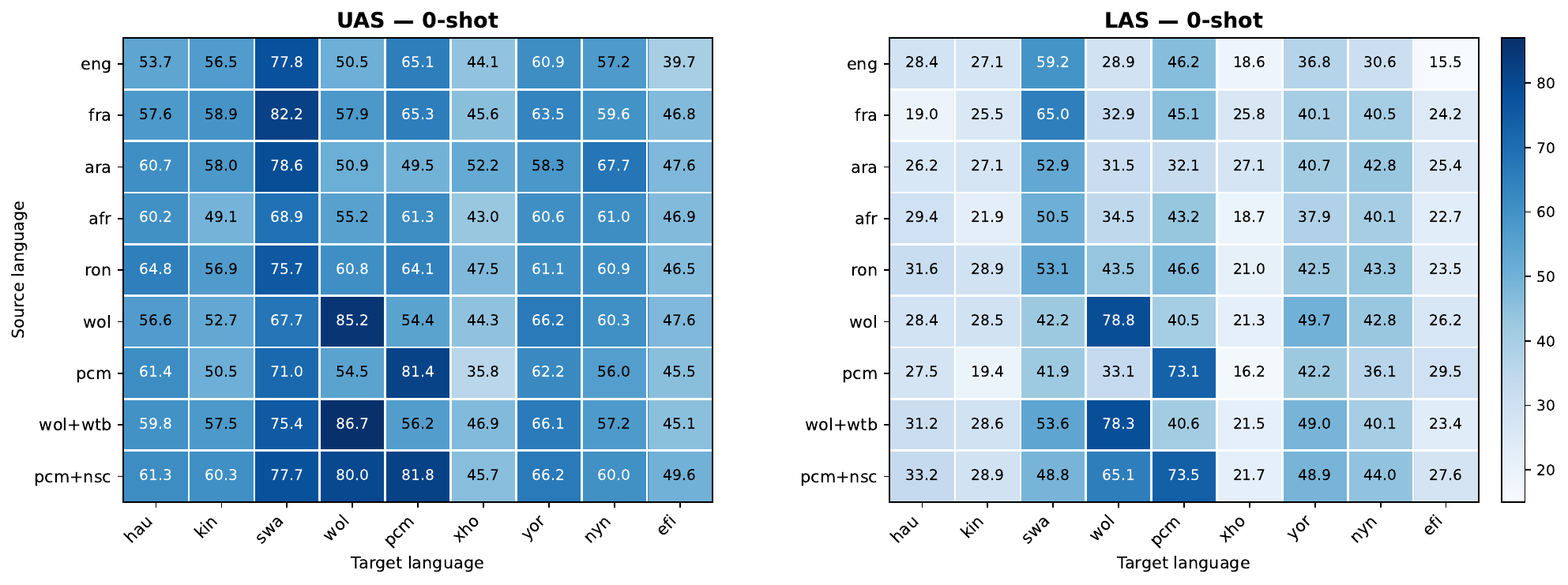}
   \caption{\textbf{Zero-shot cross-lingual transfer performance of AfroXLMR-large-76L on AfriSUD.} Rows denote the source language used for transfer and columns denote the target language. 
   }
   \label{fig:transfer_heat}
 \end{figure*}
 
Figure~\ref{fig:transfer_heat} summarizes the 0-shot transfer with AfroXLMR-large-76L across source--target pairs. \texttt{pcm+nsc} performs best overall, while \texttt{ron} is the best among non-African sources. Among targets, \texttt{swa} is consistently easier, while \texttt{xho} and \texttt{efi} remain more difficult. Results for all-shot settings are included in Appendix~\ref{sec:appendix_xlingual_all}. Figure~\ref{fig:relation_transfer}, breaks the results down by relation, and Appendix~\ref{app:crosslingual-relations} shows relation confusions and typological effects in Figure~\ref{fig:confusion-matrix} and Table~\ref{tab:african-delta} respectively.

\section{Analysis}

\label{sec:error-analysis}
\begin{figure}[t]
  \centering
  \includegraphics[width=\columnwidth]{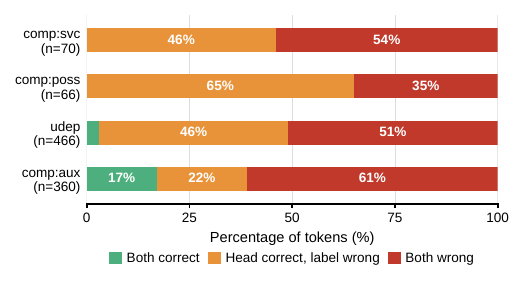}
  \caption{\textbf{Error breakdown for Gemini-3.1~Pro (0-shot) on four SUD-distinctive relations}. Bars separate fully correct predictions, label-only errors, and head-plus-label errors.}
  \label{fig:error_breakdown}
\end{figure}

\paragraph{Where Do Models Fail?} Although Gemini-3.1~Pro achieves the best overall prompting scores, the aggregate metrics do not show which constructions remain difficult. To better characterize these errors, we aggregated its 0-shot predictions across all nine African languages in AfriSUD and analyze four SUD-distinctive relations: serial-verb constructions (\texttt{comp:svc}), possessive constructions (\texttt{comp:poss}), underspecified dependencies (\texttt{udep}), and Tense-Aspect-Mood (TAM) auxiliaries (\texttt{comp:aux}). These four relations are also the most unevenly distributed across the collection (Table~\ref{tab:profile}): \texttt{comp:poss} reaches 63.0 arcs per 1K tokens in Kinyarwanda and below 4 elsewhere, and \texttt{comp:svc} is concentrated in Kinyarwanda, Runyankore, and Efik, so a
model is unlikely to recover them through cross-lingual generalization.
Figure~\ref{fig:error_breakdown} reveals a clear head--label mismatch; Gemini obtains 0\% LAS for \texttt{comp:svc} and \texttt{comp:poss}, but predicts the correct head (UAS) in 46\% and 65\% of cases respectively. For \texttt{udep}, LAS scores remains low despite the 49\% UAS of the head. The model does not produce \texttt{comp:svc} or \texttt{comp:poss} in these cases, but instead assigns broader labels such as \texttt{mod} and \texttt{comp:obj}.  \texttt{mod} accounts for 45\% of \texttt{comp:svc} errors and 14\% of \texttt{comp:poss} errors, while \texttt{comp:obj} accounts for 23\% and 42\%, respectively. On the other hand, \texttt{comp:aux} errors are more structural, head accuracy drops to 39\%, and multi-token TAM chains are often flattened. For example, Naija auxiliary chains such as \emph{don\,\ldots\,dey\,\ldots\,fit\,\ldots\,go} are predicted as flat \texttt{comp:obj} attachments rather than cascading \texttt{comp:aux} dependencies. 
Yoruba constructions involving the focus marker \emph{ni} show a similar pattern: the model promotes the main verb to \texttt{root} rather than attaching it to \emph{ni} as \texttt{comp:aux}. These errors indicate difficulty with TAM constructions beyond SUD label selection alone.
Few-shot prompting reduces some labeling errors: \texttt{comp:poss} LAS rises from 0\% to 6.8\% with one example and 7.5\% with five, while \texttt{udep} improves from 4.0\% to 11.3\% and 18.2\%. However, structurally complex relations such as \texttt{comp:aux} remain more challenging.

\begin{figure}[t]
  \centering
  \includegraphics[width=\columnwidth]{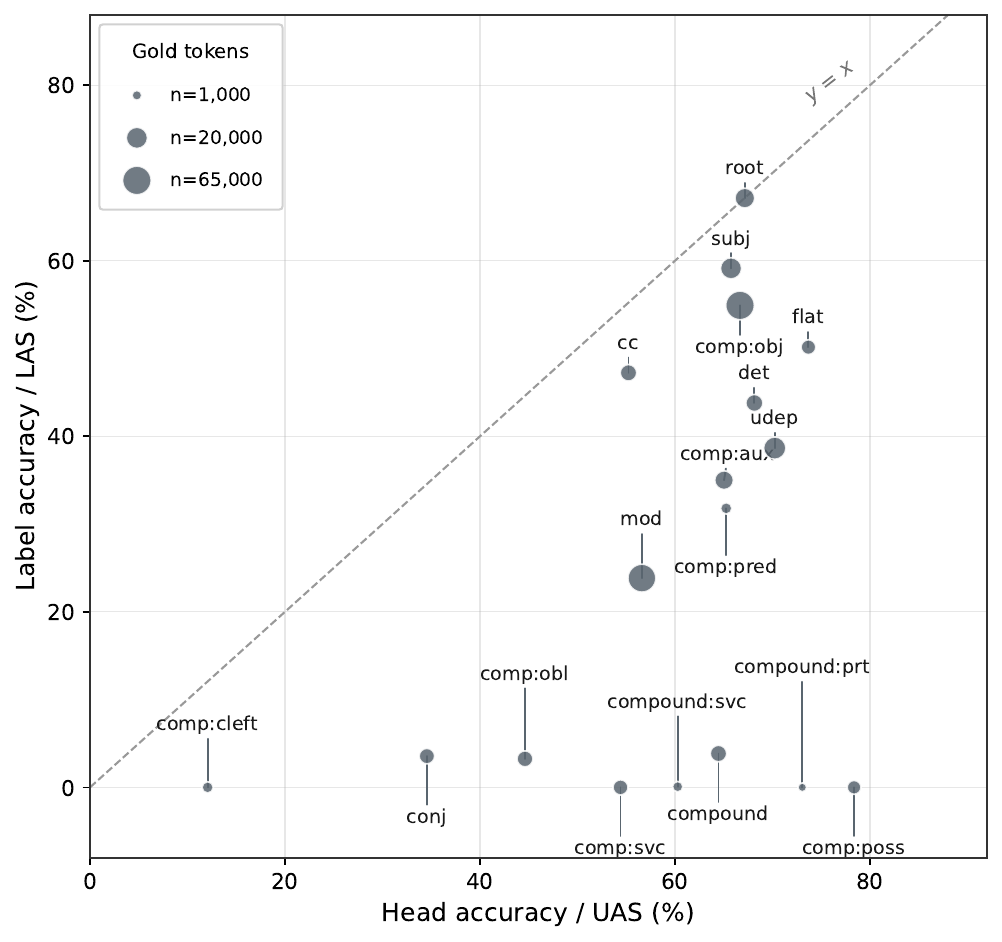}
  \caption{\textbf{AfroXLMR-large-76L UAS versus LAS accuracy for each SUD relation across 0-shot cross-lingual source→target pairs.} Point size indicates the number of gold tokens, relations far below the diagonal show a large UAS - LAS gap, indicating that the correct attachment is often identified but the dependency relation is mislabeled.}
  \label{fig:relation_transfer}
\end{figure}

\paragraph{A similar gap appears in cross-lingual transfer.} The UAS--LAS gap observed with LLM prompting, also appears in supervised cross-lingual transfer  Figure~\ref{fig:relation_transfer}. Relations such as \texttt{root}, \texttt{subj}, and \texttt{comp:obj} show comparable UAS and LAS, whereas several SUD-specific relations exhibit a much larger gap. For example, \texttt{comp:poss} and \texttt{compound:prt} exceed 70\% UAS but score 0\% LAS, with similar behavior for \texttt{comp:svc} and \texttt{compound}. This suggests that the main difficulty is to transfer relation labels rather than identify heads. 

\section{Conclusion}
\label{sec:conclusion}

We introduced AfriSUD, a coordinated SUD resource for nine African languages, and provide baseline parsers using Stanza and fine-tuned multilingual Pretrained Language Models. The results show that supervised parsers remain strong: Stanza achieves the best overall LAS, while Africa-centric PLMs such as AfroXLMR are competitive with general multilingual encoders. LLMs improve with in-context examples, but their performance remains lower than supervised parsers, especially on LAS. The persistent UAS--LAS gap shows that LLMs recover syntactic heads more reliably than SUD relation labels. Overall, AfriSUD provides a foundation for more linguistically grounded evaluation and syntactic modeling of African languages. Future work should increase annotation coverage, broaden the set of languages, constructions, and models evaluated, and extend independent double annotation across all AfriSUD languages.

\section*{Limitations} 
Our work illustrates some of the significant challenges of the annotation task, and the data cover a limited number of examples for some dependency relations. Annotation guidelines need to be refined, especially in light of some underspecified relations.
Our LLM experiments are restricted to selected models and prompting settings; closed models may change over time, making the exact reproducibility difficult. Finally, the error analysis focuses on selected SUD-particular relations and the best-performing prompting model. 

\section*{Ethics Statement or Broader Impact}
Our work is intended to support syntactic NLP research for African languages through the new AfriSUD treebanks. Most of the data come from publicly available sources except the Efik data which comes from a closed source and is therefore excluded from the public release. We do not anticipate significant privacy risks, since the released materials are based on public texts. The evaluated LLMs are used only for research purposes, and the resources we release are intended to support reproducible and inclusive NLP research.

\section*{Use of AI Assistants}
We used Claude Code (Anthropic) for debugging and developing parts of the experimental codebase. All scientific claims, experimental design, results, and conclusions were produced and verified by the authors.

\section*{Acknowledgments}
This work was supported by the Princeton Language and Intelligence (PLI) Seed Grant Program. 
\newline The authors thank the Princeton Center for Digital Humanities for its support in preparing the grant application, and the Princeton Laboratory for Artificial Intelligence for providing compute resources. We also thank the Masakhane Research Foundation for handling payments to annotators based in different parts of the world. We are grateful to Joakim Nivre and Dan Zeman for leading the initial workshop-style session on dependency relations, which helped launch the annotation training process and to Khensa Amani Daoudi for her extensive support in resolving issues with the annotation tool. Happy Buzaaba is supported by the Program in African Studies and the Africa World Initiative at Princeton. David Adelani acknowledges the support of Natural Sciences and Engineering Research Council (NSERC) of Canada, IVADO and the Canada First Research Excellence Fund, and in part by the AI2050 program at Schmidt Sciences.

\bibliography{custom}


\section{Appendix}
\appendix

\section{AfriSUD Treebank Release and Licensing Information}
\label{sec:appendix_release}
Table \ref{tab:release-status} summarizes the release and licensing information for the AfriSUD treebanks. All treebanks except Efik are publicly available with annotation guidelines and benchmark code. Efik is included in our experiments but cannot be redistributed due to restrictions on its source data. Researchers may contact the authors regarding non-commercial research access.

\begin{table}[t]
\centering
\small
\begin{tabular}{@{}llllp{4.5cm}@{}}
\toprule
\textbf{Language} & \textbf{License} & \textbf{Public} & \textbf{Restrictions} \\
\midrule
Hausa                   & CC BY-NC 4.0 & Yes & --- \\
Naija                    & CC BY-NC 4.0 & Yes & --- \\
Wolof                     & CC BY-NC 4.0 & Yes & --- \\
Yor\`ub\'a   & CC BY-NC 4.0 & Yes & --- \\
Efik                 & Closed & \textbf{No} &  No distribution \\
Swahili                  & CC BY-NC 4.0 & Yes & --- \\
Kinyarwanda              & CC BY-NC 4.0 & Yes & --- \\
Runyankore                  & CC BY-SA 4.0 & Yes & --- \\
isiXhosa                  & CC BY-NC 4.0 & Yes & --- \\
\bottomrule
\end{tabular}
\caption{\textbf{AfriSUD release status}. License, and public release status for each treebank, see Table \ref{tab:language_stats} for corpus source. Efik is excluded from the public release due to closed-source restrictions.}
\label{tab:release-status}
\end{table}

\section{An example CoNLL-U SUD annotation }
\label{sec:appendix_conllu}
\autoref{tab:conllu-example-wol} shows the CoNLL-U format of the annotated Wolof sentence. The token counts in Table\ref{tab:language_stats} include only regular CoNLL-U syntactic tokens, i.e., lines with integer IDs in Appendix Table\ref{tab:conllu-example-wol}. Multiword-token lines, empty nodes, comments, and blank lines are excluded. Thus, counts reflect the decomposed tokens used for dependency annotation rather than undecomposed surface forms.


\definecolor{posNOUN}{RGB}{166,206,227}
\definecolor{posVERB}{RGB}{178,223,138}
\definecolor{posAUX}{RGB}{253,192,134}
\definecolor{posDET}{RGB}{202,178,214}
\definecolor{posPART}{RGB}{255,255,153}
\definecolor{posPRON}{RGB}{251,154,153}
\definecolor{posPROPN}{RGB}{166,206,227}
\definecolor{posSCONJ}{RGB}{204,235,197}
\definecolor{posPUNCT}{RGB}{220,220,220}
\definecolor{headergray}{RGB}{242,242,242}

\begin{table*}[t]
\centering
\renewcommand{\arraystretch}{1.5}
\setlength{\tabcolsep}{14pt}
\begin{tabular}{@{} r l l l r l l @{}}   
\toprule
\textbf{ID} &
\textbf{Form} &
\textbf{Lemma} &
\textbf{UPOS} &
\textbf{Head} &
\textbf{Deprel} &
\textbf{Gloss} \\
\midrule
 1 & Askan & askan & \cellcolor{posNOUN}\textsc{noun}   &  3 & subj               & people  \\
 2 & wi    & bi    & \cellcolor{posDET}\textsc{det}     &  1 & det                & the     \\
 3 & dañoo & da    & \cellcolor{posAUX}\textsc{aux}     &  0 & \textbf{root}      & have    \\
 4 & war   & war   & \cellcolor{posVERB}\textsc{verb}   &  3 & comp:aux           & should  \\
 5 & a     & a     & \cellcolor{posPART}\textsc{part}   &  6 & udep               & to      \\
 6 & sàkku & sàkku & \cellcolor{posVERB}\textsc{verb}   &  3 & comp:obj           & request \\
 7 & ñu    & mu    & \cellcolor{posPRON}\textsc{pron}   &  8 & subj               & they    \\
 8 & woo   & woo   & \cellcolor{posVERB}\textsc{verb}   &  6 & comp:obj           & call    \\
 9 & Karaa & Karaa & \cellcolor{posPROPN}\textsc{propn} &  8 & comp:obj           & Karaa   \\
10 & bala  & balaa & \cellcolor{posSCONJ}\textsc{sconj} &  8 & mod                & before  \\
11 & ñoo   & a     & \cellcolor{posAUX}\textsc{aux}     & 10 & comp:obj           & they    \\
12 & diig  & diig  & \cellcolor{posVERB}\textsc{verb}   & 11 & comp:aux           & sink    \\
13 & .     &   -  & \cellcolor{posPUNCT}\textsc{punct} &  3 & punct              & .       \\
\bottomrule
\end{tabular}
\caption{\textbf{CoNLL-U/SUD annotation for the Wolof sentence}
\textit{"Askan wi dañoo war a sàkku ñu woo Karaa bala ñoo diig".}
(\textit{`The people must demand that Karaa be called in
before the team goes under.'})
UPOS tag colours are consistent with the dependency tree
visualizations throughout the paper.}
\label{tab:conllu-example-wol}
\end{table*}

\section{Dependency Structure and Relation Distributions}

\label{sec:Structural_profile}
Several sources of variation not captured by corpus size are highlighted in \autoref{tab:profile}, IsiXhosa has 69 relation labels over 7.9K tokens, compared with 29 over 19.4K tokens in Wolof and 51 over 33.3K tokens in Kinyarwanda. More importantly, 46 IsiXhosa labels and 38 Naija labels occur at most ten times, whereas only 9 Wolof labels fall into this range. This long tail of infrequent labels may make relation prediction more difficult, particularly for IsiXhosa and Efik, which show the lowest LAS across the models in Table~\ref{tab:combined_results}.

The proportion of non-projective arcs ranges from 0.01\% in Hausa to 3.63\% in IsiXhosa. IsiXhosa also has the greatest mean dependency length (3.36 tokens) and the highest proportion of long arcs, with 16.0\% spanning at least seven tokens. In contrast, Kinyarwanda has the deepest trees (12.9), but relatively short dependencies (2.11 tokens) and almost no non-projectivity. These differences show that tree depth and dependency length capture distinct aspects of structural complexity.

Following the structural differences above, selected SUD relations also show strong language-specific patterns. Serial verb relations occur at 64.6 arcs per 1,000 tokens in Kinyarwanda but only 0.2 in Wolof. Likewise, \texttt{comp:poss} occurs at 63.0 in Kinyarwanda and below 4.0 elsewhere, while \texttt{comp:cleft} reaches 108.9 in IsiXhosa and remains below 3.5 in the other treebanks, consistent with the copula less constructions discussed in \S\ref{sec:langspecific}. Some variation also reflects annotation conventions. In Swahili, for example, \texttt{udep} occurs at 186.9 per 1,000 tokens, whereas it does not occur in Yor\`ub\'a, and determination is represented with \texttt{mod} and \texttt{udep} rather than \texttt{det}. A transferred parser may therefore recover the correct heads in Swahili while still receiving lower LAS because its relation labels do not match the target annotation scheme.

\begin{table*}[p]
\centering
\footnotesize
\setlength{\tabcolsep}{4pt}
\begin{tabular}{lrrrrrrr}
\toprule
\multicolumn{8}{c}{\textbf{Inventory and structure}}\\
\midrule
 Lang & Relations $|R|$ & Labels $R_{\leq10}$ & Mean depth $Dep$ & Mean length $MDL$ & Long arcs (\%) \%$\geq$7 & \%NP$_{arcs}$ & \%NP$_{trees}$ \\
\midrule
efi & 27 & 11 & \phantom{0}4.1 & 2.39 & \phantom{0}5.8 & 0.58 & \phantom{0}2.8 \\
hau & 42 & 20 & \phantom{0}7.2 & 2.79 & 10.6 & 0.01 & \phantom{0}0.1 \\
kin & 51 & 29 & 12.9 & 2.11 & \phantom{0}5.3 & 0.01 & \phantom{0}0.1 \\
nyn & 33 & 11 & \phantom{0}4.4 & 1.95 & \phantom{0}5.3 & 0.35 & \phantom{0}1.0 \\
pcm & 65 & 38 & \phantom{0}9.8 & 2.57 & \phantom{0}6.6 & 0.41 & \phantom{0}4.2 \\
swa & 63 & 32 & \phantom{0}8.7 & 2.75 & \phantom{0}9.3 & 0.23 & \phantom{0}2.8 \\
wol & 29 & \phantom{0}9 & \phantom{0}6.9 & 2.80 & \phantom{0}9.3 & 0.95 & \phantom{0}7.1 \\
xho & 69 & 46 & \phantom{0}4.9 & 3.36 & 16.0 & 3.63 & 18.0 \\
yor & 33 & 16 & \phantom{0}4.9 & 2.61 & \phantom{0}8.7 & 1.63 & \phantom{0}8.3 \\
\midrule
\multicolumn{8}{c}{\textbf{SUD constructions, arcs per 1K tokens}}\\
\midrule
 Lang & \texttt{aux} & \texttt{svc} & \texttt{poss} & \texttt{cleft} & \texttt{udep} & \texttt{det} & \texttt{pred} \\
\midrule
efi & \phantom{0}18.5 & 22.5 & \phantom{0}0.0 & \phantom{00}1.4 & \phantom{00}1.5 & 34.3 & \phantom{0}1.1 \\
hau & 113.9 & 10.1 & \phantom{0}3.8 & \phantom{00}0.0 & \phantom{0}57.7 & 11.2 & \phantom{0}2.6 \\
kin & \phantom{00}6.4 & 64.6 & 63.0 & \phantom{00}0.0 & \phantom{0}93.9 & \phantom{0}8.4 & \phantom{0}0.5 \\
nyn & \phantom{0}26.2 & 31.1 & \phantom{0}0.0 & \phantom{00}0.0 & 153.3 & 44.9 & \phantom{0}4.8 \\
pcm & \phantom{0}54.6 & \phantom{0}3.7 & \phantom{0}0.0 & \phantom{00}3.3 & \phantom{00}0.2 & 72.2 & 14.7 \\
swa & \phantom{0}23.1 & \phantom{0}6.0 & \phantom{0}0.0 & \phantom{00}0.0 & 186.9 & \phantom{0}0.0 & \phantom{0}7.5 \\
wol & \phantom{0}61.2 & \phantom{0}0.2 & \phantom{0}0.0 & \phantom{00}1.3 & \phantom{0}85.2 & 83.0 & \phantom{0}9.0 \\
xho & \phantom{0}24.0 & 22.0 & \phantom{0}0.0 & 108.9 & \phantom{00}0.1 & 26.4 & \phantom{0}7.6 \\
yor & \phantom{0}88.5 & 18.7 & \phantom{0}0.0 & \phantom{00}0.0 & \phantom{00}0.0 & 30.6 & \phantom{0}0.5 \\
\bottomrule
\end{tabular}
\caption{\textbf{Structural profile of the AfriSUD treebanks.}
The upper panel summarizes relation inventories and dependency structure statistics, the lower panel reports selected SUD relation frequencies measured as dependency arcs per 1,000 tokens. $|R|$: distinct dependency labels, $R_{\leq10}$: labels occurring at most 10 times, $Dep$: mean tree depth, $MDL$: mean dependency length, $\%\geq7$: percentage of arcs spanning at least seven tokens, $\%NP_{\mathrm{arcs}}$ and $\%NP_{\mathrm{trees}}$: non-projective arcs and trees.
Column headers in the lower panel abbreviate the corresponding SUD relation labels: \texttt{aux} (\texttt{comp:aux}), \texttt{poss} (\texttt{comp:poss}), \texttt{cleft} (\texttt{comp:cleft}), \texttt{udep}, \texttt{det}, and \texttt{pred} (\texttt{comp:pred}). The \texttt{svc} column combines \texttt{comp:svc} and \texttt{compound:svc}, reflecting differences in annotation choices across language teams
}
\label{tab:profile}
\end{table*}


\section{Cross-lingual transfer across all source–target pairs}

\begin{figure*}[t]
  \centering
  \includegraphics[width=0.9\textwidth]{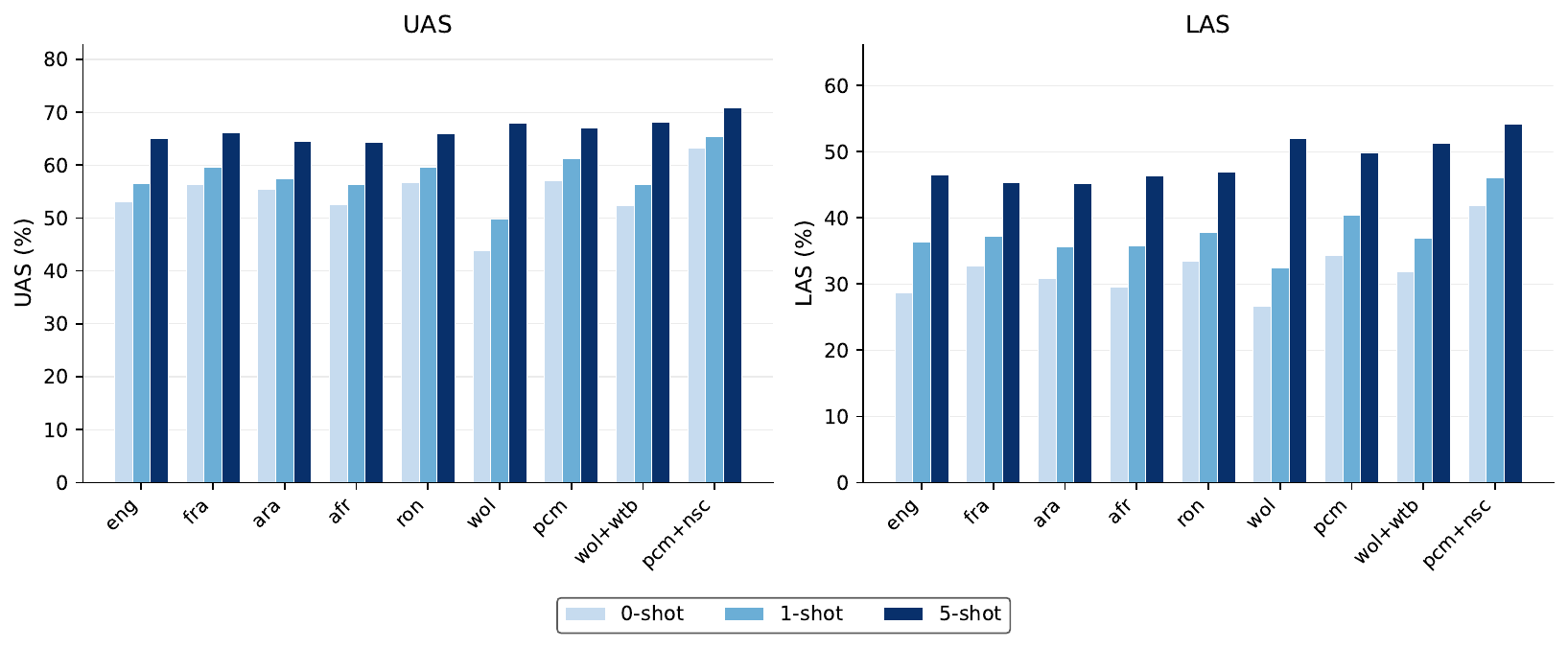}
  \caption{\textbf{Cross-lingual transfer across shot settings.} Average UAS and LAS of AfroXLMR-large-76L across source languages under 0-, 1-, and 5-shot transfer.}
  \label{fig:transfer_barcharts}
\end{figure*}

\label{sec:appendix_xlingual_all}
\autoref{tab:basexlingual} extends the 0-shot results in Figure~\ref{fig:transfer_heat} with 1- and 5-shot transfer for AfroXLMR-large-76L across all source--target pairs. \texttt{pcm+nsc} achieves the highest average UAS at all shot levels (63.3/65.4/70.8 UAS at 0/1/5-shot), while few-shot examples improve typologically distant pairs for example  LAS increases from 26.8 at 0-shot to 60.4 at 5-shot for \texttt{wol}$\rightarrow$\texttt{nyn} pair. \texttt{swa} remains the easiest target language and \texttt{efi} the hardest across all settings. Figure~\ref{fig:transfer_barcharts} shows the corresponding average UAS and LAS per source language across shot settings.
\begin{table*}[p]
\centering
\scriptsize
\setlength{\tabcolsep}{4pt}
\renewcommand{\arraystretch}{1.08}
\begin{tabular}{llrrrrrrrrrrr}
\toprule
\textbf{Metric} & \textbf{Source} & \textbf{\# shot} & \textbf{efi} & \textbf{hau} & \textbf{kin} & \textbf{nyn} & \textbf{pcm} & \textbf{swa} & \textbf{wol} & \textbf{xho} & \textbf{yor} & \textbf{Avg} \\
\midrule
\multirow{27}{*}{\textbf{UAS}}
& \multirow{3}{*}{eng}
   & 0 & 39.9 & 54.2 & 57.6 & 50.1 & 67.2 & 77.3 & 33.8 & 43.0 & 54.5 & 53.1{\tiny$\pm$12.8} \\
&  & 1 & 48.1 & 59.2 & 58.6 & 54.3 & 67.8 & 78.7 & 38.1 & 49.0 & 55.7 & 56.6{\tiny$\pm$11.1} \\
&  & 5 & 51.0 & 66.2 & 62.8 & 71.2 & 71.2 & 79.9 & 66.7 & 51.6 & 65.0 & 65.1{\tiny$\pm$8.7} \\
\cmidrule(lr){2-13}
& \multirow{3}{*}{fra}
   & 0 & 44.1 & 59.0 & 59.3 & 54.9 & 65.7 & 82.5 & 33.9 & 48.8 & 59.2 & 56.4{\tiny$\pm$13.0} \\
&  & 1 & 49.5 & 62.2 & 61.3 & 58.5 & 67.2 & 82.8 & 43.1 & 53.0 & 59.2 & 59.6{\tiny$\pm$10.7} \\
&  & 5 & 51.4 & 65.0 & 64.4 & 73.0 & 69.7 & \textbf{84.5} & 65.9 & 56.0 & 64.9 & 66.1{\tiny$\pm$9.0} \\
\cmidrule(lr){2-13}
& \multirow{3}{*}{ara}
   & 0 & 48.0 & 60.9 & 58.3 & 66.5 & 47.4 & 78.8 & 29.4 & 52.1 & 57.9 & 55.5{\tiny$\pm$13.0} \\
&  & 1 & 51.2 & 63.9 & 58.6 & 65.5 & 50.2 & 78.6 & 36.3 & 53.8 & 58.3 & 57.4{\tiny$\pm$11.1} \\
&  & 5 & 52.2 & 66.0 & 61.1 & 74.6 & 59.3 & 80.5 & 63.3 & \textbf{58.6} & 65.7 & 64.6{\tiny$\pm$8.1} \\
\cmidrule(lr){2-13}
& \multirow{3}{*}{afr}
   & 0 & 43.8 & 61.4 & 49.4 & 58.1 & 61.1 & 69.8 & 30.9 & 43.9 & 54.7 & 52.6{\tiny$\pm$11.1} \\
&  & 1 & 45.8 & 65.4 & 54.2 & 65.0 & 64.2 & 70.7 & 35.9 & 48.9 & 57.0 & 56.4{\tiny$\pm$10.6} \\
&  & 5 & 50.3 & 67.3 & 58.1 & 74.7 & 66.0 & 75.8 & 64.5 & 57.4 & 65.9 & 64.4{\tiny$\pm$7.7} \\
\cmidrule(lr){2-13}
& \multirow{3}{*}{ron}
   & 0 & 49.4 & 64.0 & 56.8 & 58.6 & 64.5 & 74.8 & 36.4 & 48.6 & 58.2 & 56.8{\tiny$\pm$10.4} \\
&  & 1 & 49.1 & 64.9 & 60.2 & 58.8 & 63.7 & 81.3 & 41.9 & 54.2 & 62.7 & 59.7{\tiny$\pm$10.4} \\
&  & 5 & 51.2 & 67.9 & 62.7 & 73.0 & 67.8 & 82.5 & 66.9 & 57.3 & 65.1 & 66.0{\tiny$\pm$8.4} \\
\cmidrule(lr){2-13}
& \multirow{3}{*}{wol}
   & 0 & 39.0 & 38.4 & 37.9 & 47.5 & 29.5 & 40.5 & 81.6 & 35.7 & 44.7 & 43.9{\tiny$\pm$14.2} \\
&  & 1 & 40.1 & 48.0 & 48.1 & 44.2 & 38.5 & 55.7 & 81.0 & 42.9 & 50.5 & 49.9{\tiny$\pm$12.1} \\
&  & 5 & \textbf{53.4} & 66.0 & 61.1 & \textbf{75.2} & 66.8 & 77.1 & 86.0 & 57.8 & 68.9 & 68.0{\tiny$\pm$9.6} \\
\cmidrule(lr){2-13}
& \multirow{3}{*}{pcm}
   & 0 & 45.9 & 62.5 & 53.8 & 61.0 & 81.0 & 72.8 & 36.4 & 40.6 & 59.3 & 57.0{\tiny$\pm$13.8} \\
&  & 1 & 50.5 & 66.2 & 60.5 & 66.9 & 81.4 & 74.5 & 36.8 & 55.5 & 59.4 & 61.3{\tiny$\pm$12.4} \\
&  & 5 & 51.8 & 70.6 & 65.4 & 72.5 & 81.4 & 76.2 & 63.0 & 55.9 & 67.1 & 67.1{\tiny$\pm$8.9} \\
\cmidrule(lr){2-13}
& \multirow{3}{*}{wol+wtb}
   & 0 & 41.7 & 49.4 & 51.3 & 50.7 & 38.9 & 59.9 & 85.0 & 45.2 & 49.1 & 52.4{\tiny$\pm$12.9} \\
&  & 1 & 42.3 & 57.8 & 58.7 & 50.0 & 43.6 & 66.6 & 84.7 & 53.6 & 49.2 & 56.3{\tiny$\pm$12.4} \\
&  & 5 & 48.7 & 66.7 & 63.3 & 74.3 & 66.3 & 81.5 & \textbf{86.6} & 56.7 & 69.3 & 68.2{\tiny$\pm$11.0} \\
\cmidrule(lr){2-13}
& \multirow{3}{*}{pcm+nsc}
   & 0 & 49.1 & 64.5 & 59.1 & 56.4 & \textbf{81.9} & 77.2 & 77.0 & 47.5 & 57.4 & 63.3{\tiny$\pm$11.9} \\
&  & 1 & 48.8 & 69.7 & 63.3 & 53.5 & 81.7 & 76.8 & 76.2 & 58.1 & 60.9 & 65.4{\tiny$\pm$10.7} \\
&  & 5 & 52.7 & \textbf{71.4} & \textbf{66.3} & 72.0 & 81.9 & 80.7 & 82.3 & 58.4 & \textbf{71.2} & \textbf{70.8{\tiny$\pm$9.8}} \\
\midrule
\multirow{27}{*}{\textbf{LAS}}
& \multirow{3}{*}{eng}
   & 0 & 17.4 & 28.7 & 27.6 & 22.7 & 48.6 & 58.9 & 6.3 & 18.0 & 30.4 & 28.7{\tiny$\pm$15.3} \\
&  & 1 & 29.9 & 32.5 & 27.5 & 39.6 & 49.0 & 62.2 & 21.2 & 30.8 & 35.3 & 36.4{\tiny$\pm$11.7} \\
&  & 5 & 33.5 & 39.9 & 37.1 & 56.5 & 52.9 & 63.7 & 55.5 & 33.6 & 46.0 & 46.5{\tiny$\pm$10.5} \\
\cmidrule(lr){2-13}
& \multirow{3}{*}{fra}
   & 0 & 23.9 & 19.8 & 25.7 & 37.4 & 45.4 & 64.7 & 11.7 & 29.1 & 36.5 & 32.7{\tiny$\pm$14.8} \\
&  & 1 & 33.6 & 25.9 & 27.6 & 41.8 & 46.6 & 65.4 & 23.1 & 35.4 & 35.4 & 37.2{\tiny$\pm$12.2} \\
&  & 5 & 33.9 & 33.2 & 34.1 & 55.6 & 51.4 & \textbf{67.3} & 52.3 & 39.7 & 40.5 & 45.3{\tiny$\pm$11.2} \\
\cmidrule(lr){2-13}
& \multirow{3}{*}{ara}
   & 0 & 25.5 & 26.1 & 27.9 & 41.2 & 29.8 & 53.3 & 9.0 & 25.7 & 39.9 & 30.9{\tiny$\pm$11.8} \\
&  & 1 & 33.7 & 30.0 & 28.8 & 44.3 & 32.0 & 57.3 & 19.6 & 32.5 & 42.8 & 35.7{\tiny$\pm$10.3} \\
&  & 5 & 36.6 & 36.5 & 34.3 & 57.0 & 44.1 & 60.1 & 51.4 & 37.7 & 48.7 & 45.2{\tiny$\pm$9.0} \\
\cmidrule(lr){2-13}
& \multirow{3}{*}{afr}
   & 0 & 23.4 & 30.0 & 22.5 & 35.8 & 42.0 & 49.6 & 8.3 & 21.5 & 32.6 & 29.5{\tiny$\pm$11.6} \\
&  & 1 & 32.3 & 36.8 & 27.0 & 43.1 & 44.4 & 53.1 & 19.3 & 32.2 & 34.3 & 35.8{\tiny$\pm$9.5} \\
&  & 5 & 34.9 & 40.2 & 34.1 & 59.3 & 49.8 & 58.6 & 53.2 & 38.8 & 48.1 & 46.3{\tiny$\pm$9.2} \\
\cmidrule(lr){2-13}
& \multirow{3}{*}{ron}
   & 0 & 27.4 & 31.0 & 29.9 & 40.6 & 47.2 & 50.2 & 15.6 & 21.3 & 38.7 & 33.5{\tiny$\pm$10.9} \\
&  & 1 & 32.9 & 33.1 & 31.1 & 38.0 & 45.7 & 61.6 & 22.2 & 32.1 & 43.9 & 37.8{\tiny$\pm$10.7} \\
&  & 5 & 32.1 & 39.6 & 38.4 & 57.7 & 50.1 & 64.4 & 55.9 & 35.6 & 49.4 & 47.0{\tiny$\pm$10.5} \\
\cmidrule(lr){2-13}
& \multirow{3}{*}{wol}
   & 0 & 19.3 & 14.1 & 18.1 & 26.8 & 19.3 & 20.4 & 74.9 & 18.4 & 28.6 & 26.7{\tiny$\pm$17.6} \\
&  & 1 & 23.8 & 24.6 & 23.5 & 27.1 & 23.6 & 40.2 & 73.5 & 26.5 & 29.3 & 32.5{\tiny$\pm$15.3} \\
&  & 5 & 36.9 & 43.0 & 40.8 & \textbf{60.4} & 52.5 & 60.9 & \textbf{79.1} & \textbf{42.7} & 51.6 & 52.0{\tiny$\pm$12.5} \\
\cmidrule(lr){2-13}
& \multirow{3}{*}{pcm}
   & 0 & 28.7 & 28.7 & 21.3 & 39.4 & 72.6 & 43.2 & 16.8 & 21.2 & 37.8 & 34.4{\tiny$\pm$16.0} \\
&  & 1 & 35.1 & 36.1 & 28.6 & 44.4 & 71.6 & 53.2 & 19.0 & 38.8 & 37.3 & 40.4{\tiny$\pm$14.2} \\
&  & 5 & \textbf{37.3} & 45.0 & 38.7 & 55.1 & 72.8 & 59.3 & 50.7 & 39.3 & 50.7 & 49.9{\tiny$\pm$10.9} \\
\cmidrule(lr){2-13}
& \multirow{3}{*}{wol+wtb}
   & 0 & 20.9 & 22.5 & 25.4 & 28.5 & 24.1 & 35.5 & 76.1 & 22.0 & 30.9 & 31.8{\tiny$\pm$16.3} \\
&  & 1 & 24.8 & 29.5 & 30.2 & 32.6 & 27.3 & 48.5 & 76.7 & 33.5 & 29.4 & 37.0{\tiny$\pm$15.4} \\
&  & 5 & 32.7 & 43.1 & 39.1 & 58.6 & 52.3 & 65.2 & 78.3 & 38.3 & 54.1 & 51.3{\tiny$\pm$13.8} \\
\cmidrule(lr){2-13}
& \multirow{3}{*}{pcm+nsc}
   & 0 & 28.9 & 33.7 & 28.5 & 37.3 & \textbf{73.8} & 48.1 & 62.8 & 23.9 & 38.9 & 41.8{\tiny$\pm$15.9} \\
&  & 1 & 32.6 & 37.4 & 32.5 & 37.4 & 73.5 & 57.0 & 66.7 & 40.1 & 37.0 & 46.0{\tiny$\pm$14.6} \\
&  & 5 & 36.5 & \textbf{47.6} & \textbf{41.9} & 55.5 & 73.5 & 63.5 & 72.5 & 40.8 & \textbf{55.7} & \textbf{54.2{\tiny$\pm$12.9}} \\
\bottomrule
\end{tabular}
\caption{\textbf{UAS and LAS for cross-lingual transfer across all source--target pairs using the base model.}
Scores are percentages. Avg reports the macro-average across languages with mean per-language standard deviation. Bold indicates the best result per metric and target language.}
\label{tab:basexlingual}
\end{table*}


\subsection{Relation Confusions and Typological Effects}
\label{app:crosslingual-relations}

Cross-lingual labeling errors are concentrated in a small set of relations rather than being uniformly distributed. As shown in Figure~\ref{fig:confusion-matrix}, \texttt{comp:aux} and \texttt{comp:svc} are frequently predicted as \texttt{comp:obj} (29\% and 53\%, respectively), while \texttt{compound} is often confused with \texttt{mod} (32\%).

Table~\ref{tab:african-delta} shows that African-language sources help some relations but not others. They improve \texttt{comp:aux} (+9.7) and \texttt{comp:obj} (+5.5), but substantially reduce accuracy on \texttt{udep} ($-$20.6), whose frequency is more closely tied to target-language annotation conventions.


\begin{table*}[t]
\centering
\begin{minipage}[c]{0.54\textwidth}
    \centering
    \includegraphics[width=\linewidth]{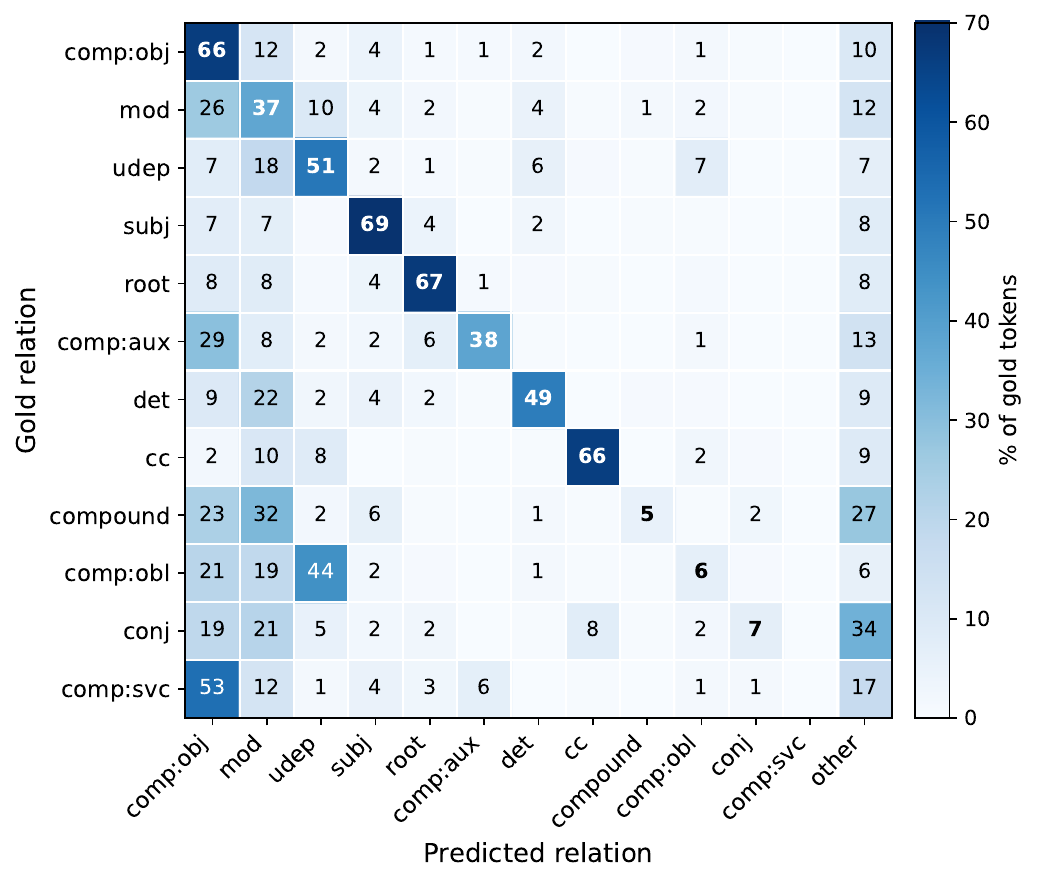}
    \captionof{figure}{Relation confusion matrix, 0-shot cross-lingual transfer, row-normalized.}
    \label{fig:confusion-matrix}
\end{minipage}
\hfill
\begin{minipage}[c]{0.42\textwidth}
    \centering
    \resizebox{\linewidth}{!}{%
    \begin{tabular}{@{}lrrr@{}}
    \toprule
    \textbf{Relation} & \textbf{Afr. src.} & \textbf{Non-Afr. src.} & \boldmath{$\Delta$} \\
    \midrule
    comp:aux   & 40.9 & 31.2 & \textbf{+9.7} \\
    conj       & 9.2  & 0.0  & \textbf{+9.2} \\
    cc         & 50.5 & 44.9 & +5.6 \\
    comp:obj   & 58.3 & 52.7 & +5.5 \\
    compound   & 6.6  & 1.7  & +4.9 \\
    comp:obl   & 4.6  & 2.4  & +2.1 \\
    subj       & 60.3 & 58.4 & +1.8 \\
    det        & 44.9 & 43.2 & +1.8 \\
    comp:svc   & 0.0  & 0.0  & 0.0 \\
    root       & 66.6 & 67.5 & $-$0.9 \\
    mod        & 19.2 & 27.1 & \textbf{$-$7.9} \\
    udep       & 26.9 & 47.5 & \textbf{$-$20.6} \\
    \bottomrule
    \end{tabular}%
    }
    \captionof{table}{LAS per relation, African-source vs.\ non-African-source transfer, 0-shot AfroXLMR-large-76L. $\Delta$ = African $-$ non-African.}
    \label{tab:african-delta}
\end{minipage}
\end{table*}

\section{POS, UAS, and LAS score for LLMS across all languages}
\label{sec:appendix_LLM_eval}

\begin{table*}[t]
\centering
\footnotesize
\setlength{\tabcolsep}{3.8pt}
\renewcommand{\arraystretch}{1.05}
\begin{tabular}{llrrrrrrrrrrrr}
\toprule
\textbf{Metric} & \textbf{Model} & \textbf{\# shot}
& \textbf{efi} & \textbf{hau} & \textbf{kin} & \textbf{nyn}
& \textbf{pcm} & \textbf{swa} & \textbf{wol} & \textbf{xho}
& \textbf{yor} & \textbf{Avg} \\
\midrule
\multirow{16}{*}{\textbf{POS}}
& \multirow{3}{*}{Gemini-3.1-Pro}
  & 0 & 79.7 & 72.3 & 79.6 & 84.2 & 85.7 & 85.7 & 88.9 & 64.9 & 77.0 & 80.2{\tiny$\pm$0.4} \\
& & 1 & 80.6 & 77.0 & 86.2 & 83.4 & 86.0 & 85.8 & 90.3 & 70.0 & 79.3 & 82.6{\tiny$\pm$0.2} \\
& & 5 & \textbf{82.7} & \textbf{84.6} & \textbf{92.0} & \textbf{84.9} & 86.4 & \textbf{87.4} & \textbf{91.8} & 81.0 & 84.8 & \textbf{86.7}{\tiny$\pm$0.3} \\
\cmidrule(lr){2-13}
& \multirow{3}{*}{GPT-5.2}
  & 0 & 55.9 & 71.3 & 76.3 & 78.5 & 85.2 & 84.4 & 82.9 & 62.3 & 77.5 & 76.4{\tiny$\pm$0.4} \\
& & 1 & 60.2 & 75.8 & 83.6 & 82.4 & 85.5 & 84.9 & 85.3 & 66.5 & 79.5 & 79.5{\tiny$\pm$0.6} \\
& & 5 & 66.0 & 82.3 & 91.0 & 83.9 & 86.5 & 86.9 & 88.3 & 75.3 & 83.5 & 83.8{\tiny$\pm$0.7} \\
\cmidrule(lr){2-13}
& \multirow{3}{*}{GPT-4o}
  & 0 & 54.0 & 72.6 & 76.8 & 76.6 & 83.9 & 71.4 & 76.0 & 63.0 & 68.8 & 72.3{\tiny$\pm$0.4} \\
& & 1 & 60.9 & 73.8 & 82.3 & 77.4 & 86.0 & 70.6 & 82.7 & 67.2 & 68.8 & 75.2{\tiny$\pm$0.3} \\
& & 5 & 64.8 & 76.7 & 88.9 & 80.5 & \textbf{87.5} & 72.7 & 87.1 & 71.8 & 70.5 & 78.6{\tiny$\pm$0.1} \\
\cmidrule(lr){2-13}
& \multirow{3}{*}{Gemma-3-12B}
  & 0 & 49.2 & 58.5 & 70.2 & 67.7 & 70.2 & 70.5 & 47.6 & 39.3 & 58.7 & 60.1{\tiny$\pm$4.3} \\
& & 1 & 49.2 & 64.0 & 78.4 & 66.5 & 72.8 & 72.9 & 52.4 & 45.3 & 64.4 & 64.3{\tiny$\pm$2.9} \\
& & 5 & 54.6 & 57.7 & 83.8 & 70.1 & 75.8 & 72.3 & 63.1 & 44.2 & 66.9 & 66.6{\tiny$\pm$3.5} \\
\cmidrule(lr){2-13}
& \multirow{3}{*}{Gemma-3-27B}
  & 0 & 52.9 & 62.5 & 72.7 & 63.7 & 79.6 & 75.4 & 49.0 & 47.1 & 63.1 & 64.4{\tiny$\pm$1.6} \\
& & 1 & 54.7 & 67.2 & 78.5 & 70.6 & 78.6 & 75.0 & 58.0 & 51.7 & 66.2 & 67.9{\tiny$\pm$2.1} \\
& & 5 & 59.8 & 72.8 & 85.1 & 74.9 & 80.0 & 75.5 & 65.8 & 62.0 & 69.8 & 72.7{\tiny$\pm$2.1} \\
\cmidrule(lr){2-13}
& Gemma-3-12B$_\textsc{sft}$
  & \textsc{ft} & 81.6 & 80.6 & 82.8 & 80.6 & 76.5 & 75.8 & 80.4 & \textbf{83.1} & \textbf{86.9} & 80.9{\tiny$\pm$3.4} \\
\midrule
\multirow{16}{*}{\textbf{UAS}}
& \multirow{3}{*}{Gemini-3.1-Pro}
  & 0 & 57.9 & 55.7 & 62.4 & \textbf{83.2} & 61.3 & 82.6 & 70.9 & 55.7 & 62.5 & 65.8{\tiny$\pm$0.3} \\
& & 1 & 57.8 & 66.4 & 63.5 & 82.9 & 68.9 & 83.4 & 77.0 & 56.3 & 67.7 & 69.7{\tiny$\pm$1.0} \\
& & 5 & 58.4 & 68.6 & 65.8 & 82.8 & \textbf{73.2} & \textbf{83.7} & \textbf{84.9} & 57.9 & \textbf{77.5} & \textbf{73.0}{\tiny$\pm$1.0} \\
\cmidrule(lr){2-13}
& \multirow{3}{*}{GPT-5.2}
  & 0 & 33.1 & 24.5 & 42.5 & 52.7 & 39.7 & 46.1 & 37.8 & 47.7 & 43.5 & 40.4{\tiny$\pm$0.5} \\
& & 1 & 39.9 & 45.2 & 53.8 & 74.7 & 56.9 & 74.6 & 51.3 & 50.7 & 59.3 & 56.5{\tiny$\pm$1.3} \\
& & 5 & 46.7 & 64.0 & 65.2 & 78.2 & 68.3 & 79.9 & 65.9 & 54.5 & 69.2 & 66.5{\tiny$\pm$0.9} \\
\cmidrule(lr){2-13}
& \multirow{3}{*}{GPT-4o}
  & 0 & 31.4 & 26.3 & 39.2 & 39.3 & 35.5 & 40.3 & 29.1 & 47.5 & 34.9 & 36.1{\tiny$\pm$0.3} \\
& & 1 & 40.7 & 33.9 & 51.0 & 53.1 & 43.5 & 59.4 & 34.0 & 52.1 & 41.6 & 46.1{\tiny$\pm$0.5} \\
& & 5 & 39.9 & 49.5 & 59.5 & 67.7 & 46.6 & 62.6 & 45.8 & 53.3 & 48.5 & 52.8{\tiny$\pm$0.2} \\
\cmidrule(lr){2-13}
& \multirow{3}{*}{Gemma-3-12B}
  & 0 & 28.8 & 22.3 & 36.1 & 43.6 & 26.0 & 45.8 & 22.6 & 30.5 & 25.9 & 31.2{\tiny$\pm$1.7} \\
& & 1 & 28.8 & 47.3 & 46.7 & 53.9 & 40.0 & 51.9 & 30.2 & 35.9 & 40.2 & 41.9{\tiny$\pm$2.1} \\
& & 5 & 31.3 & 45.5 & 58.5 & 63.0 & 48.5 & 59.3 & 39.3 & 34.5 & 50.3 & 48.2{\tiny$\pm$2.4} \\
\cmidrule(lr){2-13}
& \multirow{3}{*}{Gemma-3-27B}
  & 0 & 38.8 & 27.5 & 40.8 & 53.6 & 40.0 & 50.4 & 27.2 & 35.1 & 37.3 & 38.5{\tiny$\pm$1.5} \\
& & 1 & 40.2 & 41.0 & 47.4 & 60.1 & 45.5 & 58.9 & 31.6 & 38.1 & 43.9 & 45.0{\tiny$\pm$2.6} \\
& & 5 & 43.7 & 50.3 & 57.8 & 70.0 & 51.0 & 63.5 & 38.5 & 45.0 & 48.9 & 51.8{\tiny$\pm$2.8} \\
\cmidrule(lr){2-13}
& Gemma-3-12B$_\textsc{sft}$
  & \textsc{ft} & \textbf{58.8} & \textbf{74.7} & \textbf{69.6} & 79.0 & 61.4 & 65.3 & 58.3 & \textbf{64.6} & 70.0 & 66.9 {\tiny$\pm$7.1}\\
\midrule
\multirow{16}{*}{\textbf{LAS}}
& \multirow{3}{*}{Gemini-3.1-Pro}
  & 0 & 37.7 & 21.1 & 28.5 & 63.4 & 47.7 & 56.8 & 54.9 & 28.9 & 47.1 & 42.4{\tiny$\pm$0.7} \\
& & 1 & 39.6 & 39.8 & 40.6 & 65.8 & 56.1 & 64.2 & 66.1 & 37.7 & 54.4 & 51.9{\tiny$\pm$0.4} \\
& & 5 & 41.9 & 50.4 & 50.3 & \textbf{69.6} & \textbf{61.8} & \textbf{68.6} & \textbf{77.4} & 41.9 & \textbf{65.0} & \textbf{59.2}{\tiny$\pm$0.8} \\
\cmidrule(lr){2-13}
& \multirow{3}{*}{GPT-5.2}
  & 0 & 12.5 &  4.6 & 15.4 & 31.8 & 20.9 & 18.1 & 16.2 & 16.7 & 19.1 & 16.5{\tiny$\pm$0.7} \\
& & 1 & 23.8 & 24.3 & 33.2 & 56.1 & 45.0 & 56.6 & 41.5 & 29.1 & 46.1 & 39.9{\tiny$\pm$1.4} \\
& & 5 & 31.3 & 44.9 & 48.0 & 64.7 & 56.8 & 66.6 & 58.2 & 36.9 & 58.0 & 52.5{\tiny$\pm$1.6} \\
\cmidrule(lr){2-13}
& \multirow{3}{*}{GPT-4o}
  & 0 &  8.1 &  2.2 &  9.6 & 14.6 &  9.9 &  7.5 &  9.6 &  7.1 &  5.8 &  8.0{\tiny$\pm$0.1} \\
& & 1 & 21.9 &  8.8 & 28.9 & 36.6 & 31.4 & 44.2 & 23.4 & 30.1 & 28.4 & 28.9{\tiny$\pm$0.5} \\
& & 5 & 25.5 & 25.3 & 40.6 & 51.5 & 34.1 & 50.2 & 38.0 & 35.8 & 34.3 & 37.7{\tiny$\pm$0.3} \\
\cmidrule(lr){2-13}
& \multirow{3}{*}{Gemma-3-12B}
  & 0 &  1.8 &  1.1 &  3.0 &  8.6 &  5.4 &  1.8 &  2.0 &  2.1 &  2.9 &  2.9{\tiny$\pm$0.1} \\
& & 1 &  7.6 & 15.7 & 18.4 & 23.3 & 15.9 & 24.1 & 14.8 & 11.1 & 19.5 & 17.0{\tiny$\pm$1.2} \\
& & 5 & 16.3 & 28.3 & 39.7 & 46.6 & 35.1 & 43.5 & 28.1 & 24.5 & 39.5 & 33.9{\tiny$\pm$2.1} \\
\cmidrule(lr){2-13}
& \multirow{3}{*}{Gemma-3-27B}
  & 0 &  6.1 &  2.8 &  6.3 & 14.5 &  9.3 &  6.4 &  2.7 &  5.5 &  5.1 &  6.1{\tiny$\pm$0.2} \\
& & 1 & 12.7 & 14.1 & 15.2 & 25.9 & 18.5 & 24.4 & 10.1 &  9.1 & 16.2 & 16.2{\tiny$\pm$1.1} \\
& & 5 & 25.4 & 30.5 & 35.7 & 53.7 & 38.3 & 47.1 & 25.7 & 26.6 & 34.6 & 35.1{\tiny$\pm$2.4} \\
\cmidrule(lr){2-13}
& Gemma-3-12B$_\textsc{sft}$
  & \textsc{ft} & \textbf{47.5} & \textbf{64.9} & \textbf{60.3} & 67.1 & 53.8 & 55.7 & 52.8 & \textbf{56.2} & 63.4 & 58.0{\tiny$\pm$6.4} \\
\bottomrule
\end{tabular}
\caption{\textbf{Few-shot POS tagging and dependency parsing performance of LLMs across AfriSUD languages.}
POS tagging accuracy, UAS, and LAS are reported for prompting models (0/1/5-shot) and a supervised
fine-tuned model (\textsc{ft}). Each prompted cell is the mean over 5 runs, \textsc{ft} results are
the mean over 5 prompt templates. Bold indicates the best result per language and metric.}
\label{tab:pos_uas_las_llm}
\end{table*}

\autoref{tab:pos_uas_las_llm} reports the complete breakdown per-language. Gemini-3.1-Pro performs better at every shot level and in nearly every language. Despite its far smaller size, the fine-tuned Gemma-3-12B achieves the best UAS and LAS in four of the nine languages (efi, hau, kin, xho).

\section{POS Tags and Dependency relations}
\label{sec:appendix_pos-deprel}
\autoref{tab:annotation_inventory} shows the part-of-speech tags and dependency relations used in the annotation.
\begin{table*}[t]
\centering
\small
\setlength{\tabcolsep}{5pt}
\renewcommand{\arraystretch}{1.08}
\begin{tabular}{lll p{0.58\textwidth}}
\toprule
\textbf{Type} & \textbf{Label} & \textbf{Subtype} & \textbf{Description} \\
\midrule
\multicolumn{4}{l}{\textit{Universal POS tags}} \\
\midrule
UPOS & ADJ   & -- & adjective \\
UPOS & ADP   & -- & adposition \\
UPOS & ADV   & -- & adverb \\
UPOS & AUX   & -- & auxiliary \\
UPOS & CCONJ & -- & coordinating conjunction \\
UPOS & DET   & -- & determiner \\
UPOS & INTJ  & -- & interjection \\
UPOS & NOUN  & -- & noun \\
UPOS & NUM   & -- & numeral \\
UPOS & PART  & -- & particle \\
UPOS & PRON  & -- & pronoun \\
UPOS & PROPN & -- & proper noun \\
UPOS & PUNCT & -- & punctuation \\
UPOS & SCONJ & -- & subordinating conjunction \\
UPOS & SYM   & -- & symbol \\
UPOS & VERB  & -- & verb \\
UPOS & X     & -- & other \\
\midrule
\multicolumn{4}{l}{\textit{SUD dependency relations}} \\
\midrule
SUD & root       & --           & root of the sentence \\
SUD & subj       & --           & subject \\
SUD & comp       & comp:aux     & auxiliary complement \\
SUD & comp       & comp:obj     & object complement \\
SUD & comp       & comp:obl     & oblique complement \\
SUD & comp       & comp:pred    & predicative complement \\
SUD & comp       & comp:cleft   & cleft complement \\
SUD & mod        & --           & modifier \\
SUD & udep       & --           & underspecified dependency used for cases ambiguous between \textit{mod} and \textit{comp:obl} \\
SUD & compound   & compound     & regular compound \\
SUD & compound   & compound:prt & verb-particle compound \\
SUD & compound   & compound:svc & serial verb compound \\
SUD & appos      & --           & appositional modifier \\
SUD & conj       & --           & coordinate conjunct \\
SUD & cc         & --           & coordinating conjunction \\
SUD & flat       & --           & name or flat expression \\
SUD & fixed      & --           & fixed grammatical expression \\
SUD & dislocated & --           & dislocated element \\
SUD & punct      & --           & punctuation \\
\bottomrule
\end{tabular}
\caption{Annotation labels used in AfriSUD, including UPOS tags and SUD dependency relations.}
\label{tab:annotation_inventory}
\end{table*}

\section{LLM evaluation protocol}
We formulate Surface-Syntactic Universal Dependencies (SUD) annotation as a constrained JSON generation task. Given a target-language sentence and its gold token list (token id and surface form only), the model outputs a single JSON object containing \textsc{lemma}, \textsc{upos}, \textsc{head}, and \textsc{deprel} for each token. No English translation or interlinear gloss is provided.
\label{sec:llm_prompt_template}
\autoref{fig:llm_prompt_template} shows the exact prompts used for LLM parsing. 
\begin{figure*}[t]
\centering
\begin{minipage}{0.95\textwidth}

\begin{promptbox}{System Prompt}
\small
You are an expert linguist producing Surface-Syntactic Universal Dependencies (SUD) annotations. Return strict JSON only. Do not use markdown. Do not add extra text.
\end{promptbox}

\vspace{0.6em}

\begin{promptbox}{User Prompt}
\small
Task: Annotate the provided target-language sentence in Surface-Syntactic Universal Dependencies (SUD).

\vspace{0.4em}
Rules:
\begin{itemize}\setlength{\itemsep}{1pt}\setlength{\parskip}{0pt}\setlength{\parsep}{0pt}
    \item Use only the provided token list (id + form).
    \item Do not add, remove, merge, reorder, or split tokens.
    \item Keep punctuation tokens.
    \item Keep ids and forms exactly as provided.
    \item Provide one output token object per input token.
    \item \texttt{head} must be an integer (\texttt{root}=0).
    \item Use SUD-style surface-syntactic heads.
    \item Return exactly one JSON object matching the schema.
    \item Do not use translations or metadata not shown below.
\end{itemize}

SUD-specific guidance:
\begin{itemize}\setlength{\itemsep}{1pt}\setlength{\parskip}{0pt}\setlength{\parsep}{0pt}
    \item copulas may serve as heads where required by SUD
    \item adpositions/prepositions may serve as heads where required by SUD
    \item auxiliaries may serve as heads according to SUD conventions
\end{itemize}

Additional instruction: \texttt{"gloss"} is optional and may be omitted.
\end{promptbox}

\end{minipage}
\caption{\textbf{Prompt template used for LLM-based SUD annotation.}}
\label{fig:llm_prompt_template}
\end{figure*}

\begin{table*}[t]
\centering
\small
\setlength{\tabcolsep}{6pt}
\renewcommand{\arraystretch}{1.25}
\begin{tabular}{cp{12.5cm}}
\toprule
\# & Full instruction text \\
\midrule
1 &
\textit{You are a multilingual SUD dependency parser.}\\
& \textit{Given the sentence below, output only token-level CoNLL-U lines.}\\
& \textit{Do not output metadata lines.}\\[2pt]
& \texttt{Sentence: \{sentence\}}\\[2pt]
& \texttt{CoNLL-U token lines:}\\
\midrule
2 &
\textit{Parse the following sentence into SUD dependency format.}\\
& \textit{Return only CoNLL-U token rows (no comments, no explanations).}\\[2pt]
& \texttt{Input sentence: \{sentence\}}\\[2pt]
& \texttt{Output:}\\
\midrule
3 &
\textit{Task: produce a valid dependency parse in CoNLL-U/SUD.}\\
& \textit{Constraints: keep only token rows, exclude sent-id and text comments.}\\[2pt]
& \texttt{Sentence to parse:}\\
& \texttt{\{sentence\}}\\[2pt]
& \texttt{Token-level parse:}\\
\midrule
4 &
\textit{Generate the dependency annotation for this sentence.}\\
& \textit{Use CoNLL-U token lines and preserve all token columns.}\\
& \textit{Do not include any line that starts with `\#'.}\\[2pt]
& \texttt{Sentence:}\\
& \texttt{\{sentence\}}\\[2pt]
& \texttt{Answer:}\\
\midrule
5 &
\textit{You are given one sentence from AfriSUD.}\\
& \textit{Return the gold-style SUD parse as token lines only.}\\
& \textit{No metadata, no prose.}\\[2pt]
& \texttt{Sentence: \{sentence\}}\\[2pt]
& \texttt{SUD parse:}\\
\bottomrule
\end{tabular}
\caption{\textbf{Instruction templates used for SFT training.} All templates request token-level CoNLL-U output with no metadata.}
\label{tab:sft_templates_full}
\end{table*}

\end{document}